\documentclass{article}

\PassOptionsToPackage{numbers}{natbib}

\usepackage[final]{neurips_2025}

\usepackage[utf8]{inputenc} 
\usepackage[T1]{fontenc}    
\usepackage{hyperref}       
\usepackage{url}            
\usepackage{booktabs}       
\usepackage{amsfonts}       
\usepackage{nicefrac}       
\usepackage{microtype}      
\usepackage{xcolor}         

\usepackage{amsmath}
\usepackage{amsthm} 
\newtheorem{proposition}{Proposition}

\usepackage{graphicx} 
\usepackage{multirow}

\usepackage{float}
\usepackage{caption}
\usepackage{subcaption}
\usepackage{wrapfig}
\usepackage{algorithm}
\usepackage{algpseudocode}

\usepackage{xcolor}

\title{FairDICE: Fairness-Driven Offline Multi-Objective Reinforcement Learning}

\author{
    Woosung Kim\textsuperscript{1*} \quad
    Jinho Lee\textsuperscript{1*} \quad
    Jongmin Lee\textsuperscript{2\dag} \quad
    Byung-Jun Lee\textsuperscript{1,3\dag} \\
    \\
    \textsuperscript{1}Korea University \quad
    \textsuperscript{2}Yonsei University \quad
    \textsuperscript{3}Gauss Labs Inc. \\
    \texttt{\{wsk208,jinho0997,byungjunlee\}@korea.ac.kr} \\
    \texttt{jongminlee@yonsei.ac.kr} \\
    \textsuperscript{*}Equal contribution \quad
    \textsuperscript{\dag}Corresponding authors
}

\begin{document}

\maketitle

\begin{abstract}
Multi-objective reinforcement learning (MORL) aims to optimize policies in the presence of conflicting objectives, where linear scalarization is commonly used to reduce vector-valued returns into scalar signals. While effective for certain preferences, this approach cannot capture fairness-oriented goals such as Nash social welfare or max-min fairness, which require nonlinear and non-additive trade-offs. Although several online algorithms have been proposed for specific fairness objectives, a unified approach for optimizing nonlinear welfare criteria in the offline setting—where learning must proceed from a fixed dataset—remains unexplored. In this work, we present FairDICE, the first offline MORL framework that directly optimizes nonlinear welfare objective. FairDICE leverages distribution correction estimation to jointly account for welfare maximization and distributional regularization, enabling stable and sample-efficient learning without requiring explicit preference weights or exhaustive weight search. Across multiple offline benchmarks, FairDICE demonstrates strong fairness-aware performance compared to existing baselines.
\end{abstract}

\section{Introduction}
\label{intro}
Sequential decision-making in real-world domains often requires balancing multiple conflicting objectives, as seen in applications like autonomous driving \cite{kiran2021deep, cai2023distributional}, robotic manipulation \cite{huang2019learning, kang2023learning}, and wireless network resource allocation \cite{messikh2018towards}. Multi-objective reinforcement learning (MORL) addresses this challenge by providing a principled framework for learning policies that maximize aggregated returns over conflicting objectives. While standard MORL with linear scalarization focuses on maximizing a weighted sum of objective returns, MORL with nonlinear scalarization, or fair MORL, promotes fair outcomes through concave scalarization objectives, such as Nash social welfare \cite{moulin2004fair}.

The nonlinearity of fair MORL presents a major optimization challenge due to its nonlinear scalarization over objective returns and has been widely studied in online settings where agents learn through interaction. Some approaches maximize a lower bound—the expected scalarized return \cite{fan2022welfare, hayes2022expected}—while others focus solely on max-min fairness or use policy-gradient methods to directly optimize the original objective\cite{park2024max, agarwal2022multi}. However, fair MORL in the offline setting remains unexplored where the agent needs to learn fair policy from a fixed dataset and avoids risky or costly interaction with the environment.

Recent studies have explored offline MORL but primarily focus on linear scalarization, learning policies conditioned on fixed preference weights \cite{abels2019dynamic} to perform well along the Pareto front \cite{ basaklarpd, yang2019generalized}. However, these methods are unsuitable for fair MORL, which aims to maximize its fairness objectives (welfare) without explicitly specifying preferences. To this end, we formulate offline fair MORL problem to directly optimize the trade-off between welfare and distribution shift regularization required for offline RL. To the best of our knowledge, this is the first work to optimize a welfare-maximizing policy from a fixed dataset.

In this paper, we also show that while MORL with linear scalarization and fair MORL are fundamentally distinct problems with different solution spaces, they can be theoretically connected under offline regularization, sharing the same optimal solution. Building on this insight, we extend the DICE-RL framework, which optimizes the stationary distribution for offline policy learning, to handle the nonlinearity of fair MORL and develop FairDICE, our sample-based offline MORL algorithm. FairDICE effectively finds welfare-optimal policies in both discrete and continuous domains with minimal additional parameters, outperforming preference-conditioned baselines even with exhaustive weight search. In summary, our contributions are threefold:

\begin{itemize}
    \item A regularized offline MORL formulation that optimizes nonlinear welfare objectives while mitigating distributional shift.
    \item A theoretical connection between our formulation and linear scalarization under regularization, showing that \textbf{FairDICE} implicitly optimizes preference weights for welfare maximization.
    \item \textbf{FairDICE}, a practical, sample-based algorithm for offline welfare optimization.
\end{itemize}

\section{Related Work}
\label{related_work}
\paragraph{Multi-Objective RL and Welfare Objectives} Linear scalarization is a common approach within MORL that optimizes a weighted sum of returns \cite{lizotte2010efficient}. Nevertheless, it fails to capture complex trade-offs like fairness or risk sensitivity \cite{roijers2013survey, van2013scalarized, van2014multi, skalse2023limitations}. This limitation has led to growing interest in nonlinear scalarization, which enables more expressive preference modeling. Recent theoretical work has shown that it is tractable to optimize nonlinear scalarizations under smooth, concave utility functions \cite{peng2023multi, agarwal2022multi}. Such scalarization functions, including Nash social welfare and Gini indices, are optimized via online interactions \cite{fan2022welfare, siddique2023fairness}. Max-min objectives, another form of fairness-aware scalarization, have been addressed in model-free online settings with entropy regularization \cite{park2024max}. 

\paragraph{Offline RL and Offline MORL}
Offline reinforcement learning aims to learn policies from fixed datasets without further environment interaction, avoiding costly or risky exploration. A major challenge in this setting is distribution shift: deviations from the behavior policy can lead to inaccurate value estimates and suboptimal performance. To address this, various strategies have been proposed, including conservative value estimation \cite{kumar2020conservative}, divergence-regularized optimization \cite{lee2021optidice, gulcehre2021regularized}, and return-conditioned sequence modeling \cite{chen2021decision}. In the multi-objective setting, most offline approaches assume linear scalarization and require explicit preference conditioning during training or evaluation \cite{wu2021offline, lin2024policy}. However, direct optimization of nonlinear objectives in offline MORL remains largely underexplored.

\section{Preliminaries}
\label{preliminaries}
\subsection{Markov Decision Process (MDP) and Multi-objective RL}
A Markov Decision Process (MDP) is defined by the tuple $(\mathcal{S}, \mathcal{A}, T, r, p_0, \gamma)$, where $\mathcal{S}$ and $\mathcal{A}$ are the state and action spaces, $T$ is the transition probability, $r$ is the reward function, $p_0$ is the initial state distribution, and $\gamma \in [0,1)$ is the discount factor.
For a policy $\pi$, the stationary distribution $d_\pi(s,a)$ represents the discounted visitation frequency of state-action pairs and satisfies the Bellman flow constraint:
\begin{align}
    d_{\pi}(s,a)=\pi(a|s)\left((1-\gamma)p_0(s)+\gamma \sum_{\bar{s},\bar{a}} T(s|\bar{s},\bar{a})d_{\pi}(\bar{s},\bar{a})\right), \ \forall s,a
\end{align}
The expected return of policy $\pi$, defined as the discounted sum of rewards, can be represented as $J(\pi):=\sum_{s,a}d_{\pi}(s,a)r(s,a)$, and standard RL aims to find $\pi$ that maximizes $J(\pi)$.

In Multi-Objective Reinforcement Learning (MORL), the return is represented as a vector $[J_1(\pi), \dots, J_I(\pi)]$, where each $J_i(\pi)$ is defined under a corresponding reward function $r_i(s, a)$. MORL methods typically scalarize the return vector to optimize a scalar objective of the form $\sum_{i} u_i(J_i(\pi))$. Linear scalarization applies fixed weights $w_i$, reducing the problem to single-objective RL with the aggregated reward $r(s, a) = \sum_{i} w_i r_i(s, a)$. This approach focuses on recovering optimal policies across different weight configurations.

However linear scalarization does not consider fairness across objectives. To address this, we use strictly concave scalarization functions, referred to as \emph{Fair MORL}, which maximize the welfare $\sum_i u_i(J_i(\pi))$, as in Nash Social Welfare with $u_i(x) = \log(x)$. However, this nonlinearity breaks the Bellman recursion, making Temporal Difference (TD) targets difficult to define and thus unsuitable for standard value-based RL methods.

\subsection{Offline Reinforcement Learning and DICE-RL framework} 
In offline reinforcement learning (RL), an agent learns a policy from a fixed dataset $D = {\{s_i,a_i,r_i,s'_{i}\}}_{i=1}^N$, without further environment interaction, making it well-suited for scenarios where exploration is costly or unsafe. A core challenge in offline RL is distribution shift, where the learned policy deviates from the behavior policy to improve performance. Excessive deviation leads to significant off-policy estimation errors and degrades real-world performance.

To address this, the DICE-RL framework \cite{lee2021optidice} regularizes policy optimization using an $f$-divergence between the stationary distribution of the learned policy $d$ and the empirical distribution $d_D$, solving:
 \begin{align}
\max_{d \geq 0} \ & \sum_{s,a} d(s,a) r(s,a) - \beta \sum_{s,a} d_D(s,a) f\left(\frac{d(s,a)}{d_D(s,a)}\right)\nonumber \\
\text{s.t.} \ & \sum_{a}d(s,a) = (1-\gamma)p_0(s) + \gamma\sum_{\bar{s},\bar{a}}T(s|\bar{s},\bar{a})d(\bar{s},\bar{a}), \quad \forall s \label{const:Bellman}
\end{align}
where the marginalized Bellman flow constraint \eqref{const:Bellman} ensures the optimized stationary distribution $d$ is valid. The hyperparameter $\beta>0$ controls the trade-off between return maximization and distribution shift regularization. In finite domains, the optimal policy $\pi^{*}$ is recovered from $d^{*}$ via $\pi^{*}(a|s)=\frac{d^{*}(s,a)}{\sum_{a}d^{*}(s,a)} \ \forall s,a$, with policy extraction methods used in continuous domains. Full derivations appear in Appendix~\ref{Appendix:dicerlframework}.

\section{Fair MORL as Convex Optimization}
\label{formulation}
We present a convex optimization formulation for Fair MORL. Before extending it to a practical offline algorithm, we analyze its structure and reveal a connection to the Eisenberg–Gale program~\citep{devanur2008market}. We reformulate the objective to facilitate future sample-based methods and highlight key differences from linear scalarization. 

\subsection{Convex Formulation via stationary distribution}
We introduce a convex formulation of Fair MORL, where the objective returns 
$J_{i}(\pi)$ are expressed in terms of the stationary distribution. This formulation extends the dual of V-LP~\citep{nachum2020reinforcement}, the linear programming formulation of RL, by replacing its linear objective with a welfare function, a similar approach explored in~\citep{park2024max, agarwal2022multi}. Specifically, we consider the following problem, where each $u_i(x)$ is strictly concave:
\begin{align}
    \text{(P1): } \max_{d\geq0}\;&\sum_{i}u_{i}\bigg(\sum_{s,a}d(s,a) r_i(s,a)\bigg) \nonumber \\ \text{s.t.}\;&\mathcal{F}_d(s)=0, \quad \forall s \nonumber
\end{align}
where $\mathcal{F}_d(s)= (1-\gamma)p_0(s)+\gamma\sum_{\bar{s},\bar{a}}T(s|\bar{s},\bar{a})d(\bar{s},\bar{a})-\sum_{a}d(s,a)$ denotes the violation degree of the Bellman flow constraint \eqref{const:Bellman}.

 Uniqueness of the solution to (P1) follows from the strict concavity of the objective and the convex, non-empty feasible set of valid stationary distributions. When $u_{i}(x) = \log(x) \ \forall i$, the optimization takes a form similar to the Eisenberg–Gale convex program, a classical model for computing fair and Pareto-efficient market equilibria:
\begin{align}
\max_{x_{ij}\geq0}\;&\sum_{i}\log\bigg(\sum_{j}u_{ij}x_{ij}\bigg) \quad
    \text{s.t.}\;\sum_{i}x_{ij}\leq1 \ \forall j \nonumber
\end{align}
where $x_{ij}$ denotes the allocation of good $j$ to buyer $i$, and $u_{ij}$ is the utility of buyer $i$. In our setting, this corresponds to allocating actions in a single-state MDP to maximize Nash social welfare, with $i$ representing actions and $j$ representing objectives. With additional constraint \eqref{const:Bellman}, (P1) can be viewed as a sequential version of the Eisenberg–Gale program, extending the allocation problem to sequential decision-making.
\subsection{Reformulation for sample-based optimization}
Problem (P1) involves an expectation inside a concave function, which complicates the future derivation of our sample-based optimization method. To address this, we reformulate (P1) as (P2) by introducing slack variables, moving the expectation outside the concave function.
\begin{align*}
    \text{(P2): } \max_{d\geq0, k_{i}}\;&\sum_{i}u_{i}\left(k_{i}\right) \quad
    \text{s.t.}\;\sum_{s,a}d(s,a) r_i(s,a) = k_{i} \ \forall i, \quad \mathcal{F}_d(s)=0, \ \forall s
\end{align*}
where $k_{i}$ is a slack variable representing the expected return for objective $i$. The Lagrangian dual of (P2) introduces Lagrange multipliers $\mu_{i}$ for the return constraints and $\nu(s)$ for the Bellman flow constraints. The dual formulation is given by:
\begin{align}
\max_{d\geq 0, k}\min_{\nu, \mu} \ &\sum_{i}u_{i}(k_{i})+ \sum_{i}\mu_{i}\bigg(\sum_{s,a}d(s,a)r_{i}(s,a)-k_{i}\bigg) + \sum_{s}\nu(s)\mathcal{F}_{d}(s)\label{Lagrangian}
\end{align}

In the dual function \eqref{Lagrangian}, the Lagrange multiplier $\mu_i$ modulates each objective’s return, resembling the role of preference weights in Linear MORL. At optimality, the following relationship holds:
\begin{align*}
    \mu^*_{i} = u_{i}'(k^*_{i}) = u_{i}'\bigg(\sum_{s,a}d^{*}(s,a)r_{i}(s,a)\bigg)
\end{align*}
Since $u'_i(x)$ is decreasing due to the strict concavity of $u_{i}$, $\mu_{i}$ acts as an implicit preference weight that penalizes large returns. For example, when $u_{i}(x)=\log(x)$, this yields $\mu^{*}_{i}$ that corresponds to the reciprocal of $i$th return, assigning higher weight to objectives with lower returns and promoting a more balanced, fair allocation—consistent with the goal of Nash social welfare and Fair MORL.

However, although $\mu_{i}$ plays a role similar to preference weights in Linear MORL, fixing $\mu^{*}_{i}$ in Linear MORL (P3) leads to a fundamentally different optimization problem from (P2) learning $\mu_{i}$ as part of the nonlinear welfare objective. (P3) is defined below, and a simple counterexample are provided in Appendix~\ref{Appendix:counter}.
\begin{align}
    \text{(P3): } \max_{d\geq0}\;&\sum_{i}\mu^{*}_{i}\sum_{s,a}d(s,a) r_i(s,a) \quad
    \text{s.t.}\;\mathcal{F}_d(s)=0, \quad \forall s \nonumber
\end{align}
Unlike (P1), which has a unique solution by strict concavity, (P3) may have multiple optimal solutions with different welfare outcomes. Thus, the welfare-maximizing policy of Fair MORL cannot generally be found by simply sweeping over weight vectors in Linear MORL.

\section{FairDICE: Welfare Optimization for Offline Fair MORL}
\label{framework}
In this section, we introduce the Regularized Welfare Optimization framework and a corresponding sample-based algorithm for effective welfare optimization in the offline setting. Building on the DICE-RL framework applied to (P2), our method, FairDICE, directly optimizes implicit preference weights to maximize welfare. We further provide theoretical support by recovering an equivalent Regularized Linear MORL formulation.
\subsection{Regularized Welfare Optimization framework}
\label{section5.1}
We formulate our framework by incorporating an $f$-divergence between the optimized stationary distribution $d$ and the empirical data distribution $d_D$ into (P2). The trade-off between the welfare and distributional shift is controlled by a hyperparameter $\beta>0$, and $f$ is assumed to be strictly convex with $f(1)=0$. The resulting convex optimization is:
\begin{align}
    \text{(P2-reg): } \max_{d\geq0, k}\;&\sum_{i}u_{i}\left(k_i\right)-\beta\sum_{s,a}d_D(s,a)f\left(\frac{d(s,a)}{d_D(s,a)}\right) \nonumber\\
    \text{s.t.}\;&\sum_{s,a}d(s,a) r_i(s,a) = k_{i} \ \forall i, \quad \mathcal{F}_d(s)=0, \ \forall s \nonumber
\end{align}
Following the DICE-RL framework, we derive a sample-based optimization method from the Lagrangian dual of (P2-reg). We also highlight the challenges of extending (P1) directly to sample-based optimization and show how (P2) circumvents the issues. The Lagrangian is expressed as:
\begin{align}
\max_{d\geq 0, k}\min_{\nu, \mu} \ L(\nu,\mu,d,k) := &\sum_{i}\mu_{i}\bigg(\sum_{s,a}d(s,a)r_i(s,a)-k_i\bigg) -\beta \sum_{s,a}d_D(s,a) f\left(\frac{d(s,a)}{d_D(s,a)}\right) \nonumber \\
&+ \sum_{i}u_{i}\left(k_i\right) +\sum_s \nu(s)\mathcal{F}_d(s) \nonumber
\end{align}
We reparameterize the stationary distribution as $d(s,a)=w(s,a)d_D(s,a)$ and express the dual in terms of the importance weights $w$, using the identity $\sum_{s}\nu(s)\sum_{\bar{s},\bar{a}}T(s|\bar{s},\bar{a})d(\bar{s},\bar{a}) = \sum_{s,a}d(s,a)\sum_{s'}T(s'|s,a)\nu(s')$. This yields the following optimization problem:
\begin{align}
&\max_{w\geq 0, k}\min_{\nu, \mu} \ \mathbb{E}_{s\sim p_0}[(1-\gamma)\nu(s)]+\mathbb{E}_{ d_D}\left[w(s,a)e_{\nu,\mu}(s,a)-\beta f\left(w(s,a)\right)\right]-\sum_{i}\left(\mu_{i}k_{i}+u_{i}(k_{i})\right)\nonumber
\end{align}
where $e_{\nu,\mu}(s,a)=\sum_{i}\mu_{i}r_{i}(s,a)+\gamma\sum_{s'}T(s'|s,a)\nu(s')-\nu(s) \ \forall s,a$.

Applying the Lagrangian dual directly to the regularized (P1) retains expected returns inside the concave functions $u_{i}(\cdot)$, preventing direct use of importance sampling. Moreover, a naive estimator such as $\sum_{s,a}d_D(s,a)\sum_{i}u_{i}(w(s,a)r(s,a))$ introduces bias, violating the validity of importance-weighted estimation.

We further simplify the optimization by reducing parameters. Using strong duality of (P2-reg), we switch the optimization order to $\min_{\nu,\mu}\max_{w,k}$ and derive closed-form solutions for $w$ and $k_{i}$ from first-order conditions:
\begin{align}
    w^{*}_{\nu,\mu}(s,a) = \max\left(0,(f')^{-1}\left(\frac{e_{\nu,\mu}(s,a)}{\beta}\right)\right) \ \forall s,a, \quad k^{*}_{i,\mu} =(u_{i}')^{-1}(\mu_{i}) \ \forall i \nonumber
\end{align}
Substituting the closed-form solutions into the Lagrangian dual yields the final optimization, which defines the loss function of our offline algorithm, FairDICE (Fair MORL via Stationary Distribution Correction):
\begin{align}
\min_{\nu, \mu} \ \mathbb{E}_{s\sim p_0}[(1-\gamma)\nu(s)]+\mathbb{E}_{(s,a)\sim d_D}\left[\beta f^{*}_{0}\left(\frac{e_{\nu,\mu}(s,a)}{\beta}\right)\right]+\sum_{i}u^{*}_{i}(-\mu_{i}) \label{eq:FairDICE}
\end{align} 
where $f^{*}(y):=\max_{x\geq0}xy-f(x)$ and $u^{*}_{i}(y):=\max_{x}xy+u_{i}(x)$ are convex conjugate functions. Solving \eqref{eq:FairDICE} gives the optimal stationary distribution $d^{*}(s,a)=w^{*}_{\nu^{*},\mu^{*}}(s,a)d_D(s,a)$. In finite domains, the optimal policy is directly recovered via $\pi^{*}(a|s)=d^{*}(s,a)/\sum_{a}d^{*}(s,a) \ \forall s,a$.

Compared to its single-objective counterpart, OptiDICE, FairDICE introduces only one additional scalar parameter per objective, incurring minimal overhead and scaling efficiently with the number of objectives. Moreover, FairDICE extends naturally to large and continuous domains by approximating $\nu$, $\mu$ and $\pi$ with function approximators. In continuous domains, we adopt weighted behavior cloning to extract the policy from the optimal stationary distribution using the following loss:
\begin{align*}
    \max_{\pi_{\phi}}\; \mathbb{E}_{(s,a)\sim d_D}\left[w^{*}_{\nu^{*},\mu^{*}}(s,a)\log \pi_{\phi}(a|s)\right]
\end{align*}
Further details in experimental settings and algorithmic descriptions for continuous domains are provided in Appendix~\ref{appendix:implementation}.
\subsection{Equivalence to Regularized Linear MORL}
\label{sec:eqlinear}
Previously, we showed that unregularized linear and fair MORL converge to different solutions. We now demonstrate that their regularized forms converge to the same unique solution. This equivalence supports that FairDICE implicitly optimizes preference weights corresponding to those in linear MORL to directly maximize target welfare.

To formalize this, we introduce (P3-reg), an extension of DICE-RL framework to Linear MORL. While any preference weight can be used, we set it to the optimal dual variable $\mu^{*}$ from \eqref{eq:FairDICE} for analysis:
\begin{align}
    \text{(P3-reg): } \max_{d\geq0}\;&\sum_{s,a}d(s,a) \sum_{i}\mu^{*}_{i}r_i(s,a)-\beta\sum_{s,a}d_D(s,a)f\left(\frac{d(s,a)}{d_D(s,a)}\right) \nonumber \\
    \text{s.t.}\;&\mathcal{F}_d(s)=0, \quad \forall s \nonumber
\end{align}  
\begin{proposition}[Equivalence between the regularized problems]
\label{prop1}
    Let $\mu_{i}^{*}$ be the optimal multipliers obtained from \eqref{eq:FairDICE}. Then, the optimal solutions of FairDICE and (P3-reg) with $\mu_{i}^{*}$ yield the same unique optimal policy. (Proof in Appendix \ref{Appendix:propositionproof})
\end{proposition}
This equivalence suggests that FairDICE reduces to an offline Linear MORL algorithm when its $\mu$ is fixed to the preference weights. We refer to this special case as FairDICE-fixed and leverage it to optimize utilitarian welfare—the sum of objective returns—by setting all $\mu$s equal.

The equivalence also implies that sweeping over preference weights in regularized Linear MORL can recover the optimal policy found by FairDICE. However, this requires training policies across a wide range of weights and selecting the one that achieves the highest welfare, which becomes impractical as the number of objectives increases. In contrast, FairDICE effectively and efficiently optimizes implicit preference weights, directly producing an offline MORL policy that maximizes welfare.



\section{Empirical Behaviors of FairDICE}
\label{empirical}
In this section, we empirically validate our theoretical insights using a multi-objective adaptation of the classic Four-Room environment~\cite{lee2021optidice, sutton1999between} and Random MDP~\cite{lee2021optidice, laroche2019safe} as a toy example. The visualization shows how FairDICE effectively optimizes the trade-off between welfare and distribution shift (Section \ref{section5.1}) and aligns with Regularized Linear MORL while optimizing its implicit preference weight for offline welfare optimization (Section \ref{sec:eqlinear}).

\begin{figure}[t]
  \centering
  \begin{subfigure}[b]{0.23\textwidth}
    \centering
    \includegraphics[width=\linewidth]{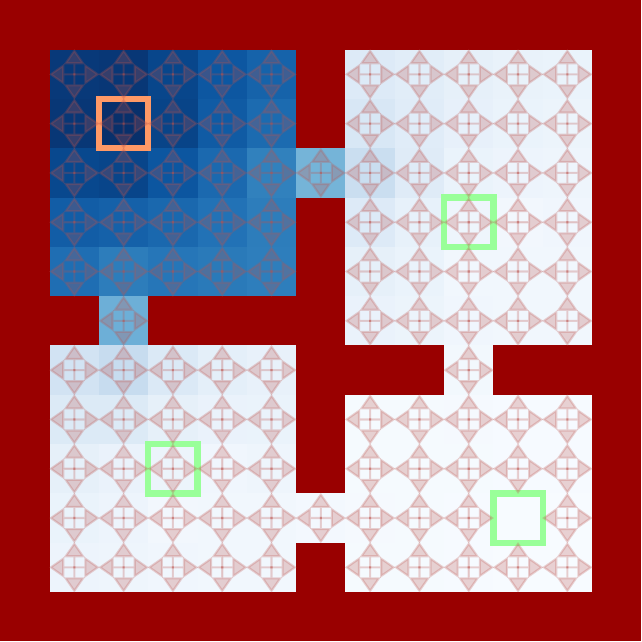}
    \caption{Data collection policy}
    \label{fig1:sub1}
  \end{subfigure}
  \hspace{0.01\textwidth}
  \begin{subfigure}[b]{0.23\textwidth}
    \centering
    \includegraphics[width=\linewidth]{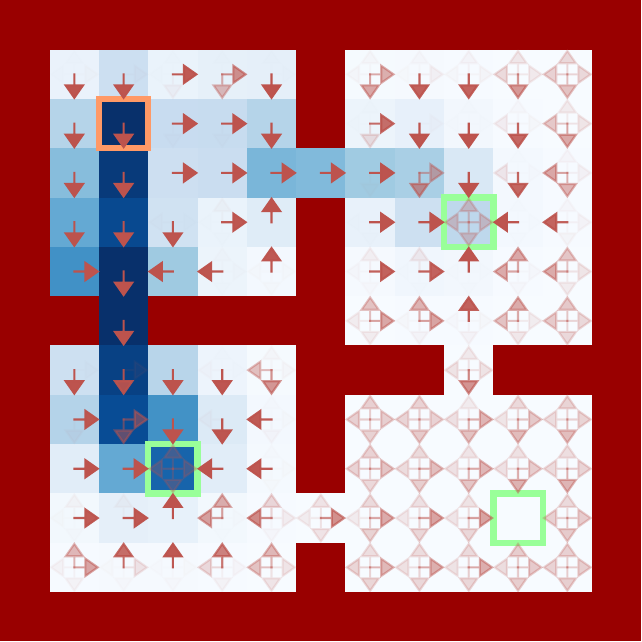}
    \caption{Utilitarian welfare}
    \label{fig:sub2}
  \end{subfigure}
  \hspace{0.01\textwidth}
  \begin{subfigure}[b]{0.23\textwidth}
    \centering
    \includegraphics[width=\linewidth]{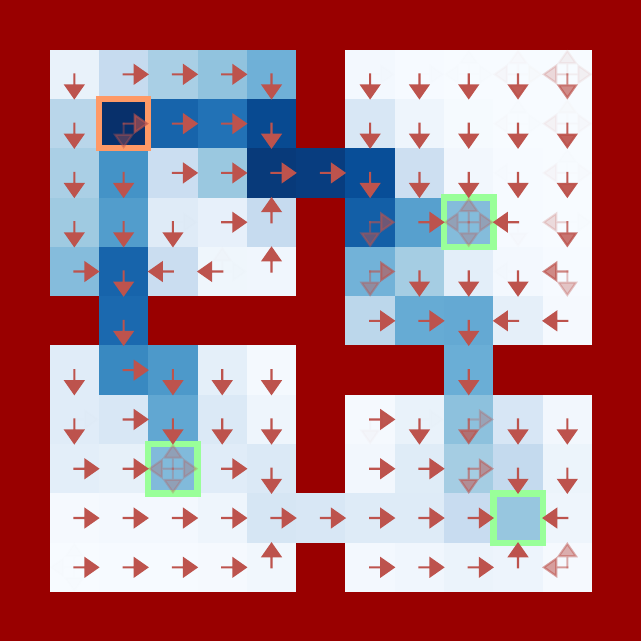}
    \caption{Nash social welfare}
    \label{fig:sub3}
  \end{subfigure}
  \hspace{0.01\textwidth}
  \begin{subfigure}[b]{0.23\textwidth}
    \centering
    \includegraphics[width=\linewidth]{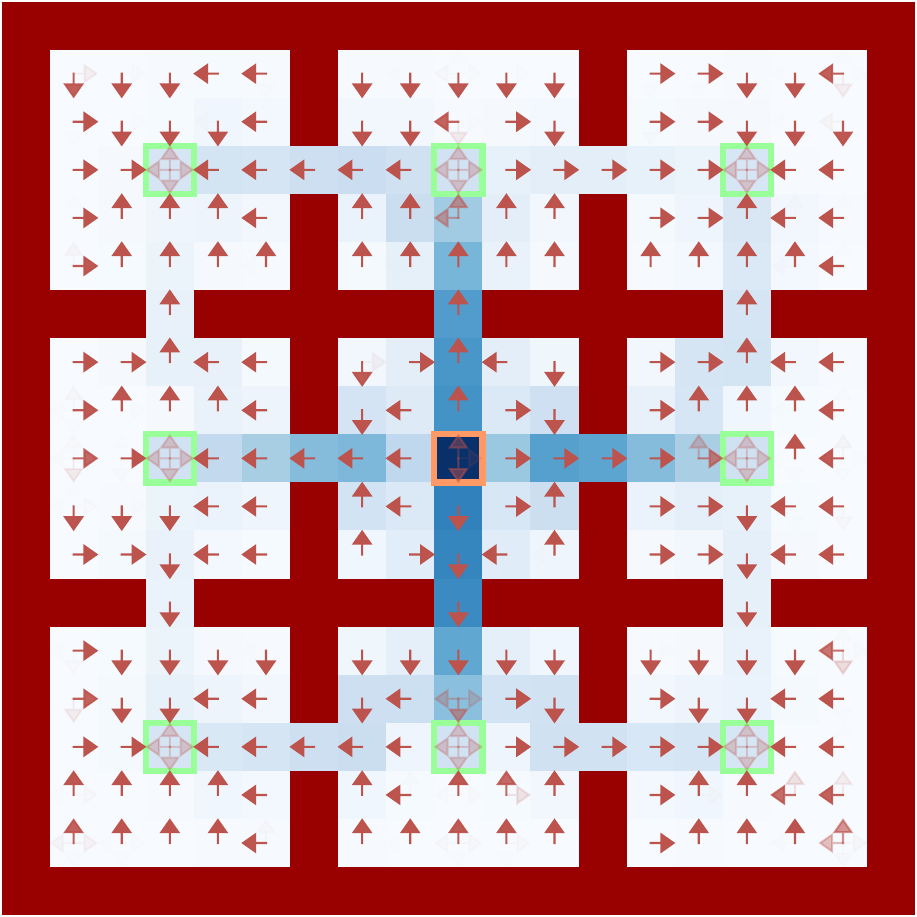}
    \caption{NSW (8 objectives)}
    \label{fig:sub4}
  \end{subfigure}
  \caption{Visualization of FairDICE policies in MO-Four-Rooms: (a) Uniformly random policy for data collection, (b) FairDICE policy maximizing Utilitarian welfare (sum of returns), (c) FairDICE policy maximizing Nash social welfare (NSW). (d) FairDICE maximizing NSW in a domain with eight objectives. Red arrows indicate the policy, and the blue heatmap shows state visitation.
}
  \label{fig1:Fourroom}
  \vspace{-0.4cm}
\end{figure}

In the experiments, we use $\alpha$-fairness to aggregate objectives, a generalized social welfare function that balances total return and fairness. The trade-off is controlled by the parameter $\alpha$: as $\alpha\to0$,  it approximates Utilitarian welfare, the sum of returns; at $\alpha=1$, it recovers Nash social welfare; and as $\alpha\to \infty$, it approaches the max-min fairness. The scalarization function is defined as:
\begin{align*}
    u_i(x) =
    \begin{cases}
        (1-\alpha)^{-1}x^{1-\alpha} \ &(\alpha \neq 1)\\
        \log(x) \ &(\alpha = 1)
    \end{cases}
     \quad \forall i
\end{align*}
We evaluate the resulting policies using three metrics: \textbf{Utilitarian welfare} $\sum_{i=1}^n R_i$, \textbf{Jain's Fairness Index} $\left( \sum_{i=1}^n R_i \right)^2 / (n \sum_{i=1}^n R_i^2)$ and \textbf{Nash social welfare (NSW)} $\sum_{i=1}^n \log(R_i)$. Utilitarian welfare measures the total return, Jain’s Fairness Index evaluates fairness across objectives, and Nash social welfare captures a trade-off between efficiency and fairness. Details of the environments and experiments introduced in this section are provided in Appendix~\ref{Appendix:Finite}.

\subsection{MO-Four-Room Experiment}
We extend the Four-Room domain to a MORL setting, referred to as MO-Four-Room, by introducing three distinct goals, each associated with a separate objective. As shown in Figure \ref{fig1:Fourroom}, the agent starts from the initial state (orange) and moves toward the goal states (green). Upon reaching a goal, it receives a one-hot reward: $[1, 0, 0]$, $[0, 1, 0]$, or $[0, 0, 1]$. To simulate offline RL, we construct a dataset of 300 trajectories collected from a uniformly random behavior policy.

\begin{wraptable}{r}{0.55\textwidth}
\centering
\caption{MO-Four-Room Performance Table}
\label{tab:nsw_comparison}
\begin{tabular}{lccc}
\toprule
         & Behavior & Utilitarian & FairDICE-NSW \\
\midrule
NSW      & -16.10   & -17.33      & \textbf{-10.77}       \\
Util     & 0.024   & \textbf{0.086}       & 0.082        \\
Jain     & 0.710   & 0.500       & \textbf{0.996}        \\
\bottomrule
\end{tabular}
\vspace{-0.35cm}
\end{wraptable}
Even when data is collected under an unfair, suboptimal policy, FairDICE successfully learns offline MORL policies that maximize target welfare. Figure~\ref{fig1:Fourroom} shows how varying $\alpha$-fairness objectives shape FairDICE’s behavior. The utilitarian objective favors the nearest goal as it results in the best return, achieving the highest sum of returns despite being unfair. However, FairDICE maximizing Nash social welfare (FairDICE-NSW), encourages balanced goal visitation, even when the environment is expanded to include eight objectives, as shown in Figure~\ref{fig:sub4}.

\subsection{Effective Optimization over Two Distinct Trade-offs}
\label{subsection_6.2}
The Regularized Welfare Optimization framework balances two trade-offs controlled by $\alpha$ and $\beta$: one between objective returns under $\alpha$-fairness and the other between welfare and distributional shift. To empirically demonstrate how these trade-offs evolve with varying parameters and generalize across different MOMDP environments, we extend the Random MDP to a MORL setting, called Random MOMDP, following a similar approach used for the Four-Room domain.

\begin{figure}[t]
  \centering
  \includegraphics[width=0.99\linewidth]{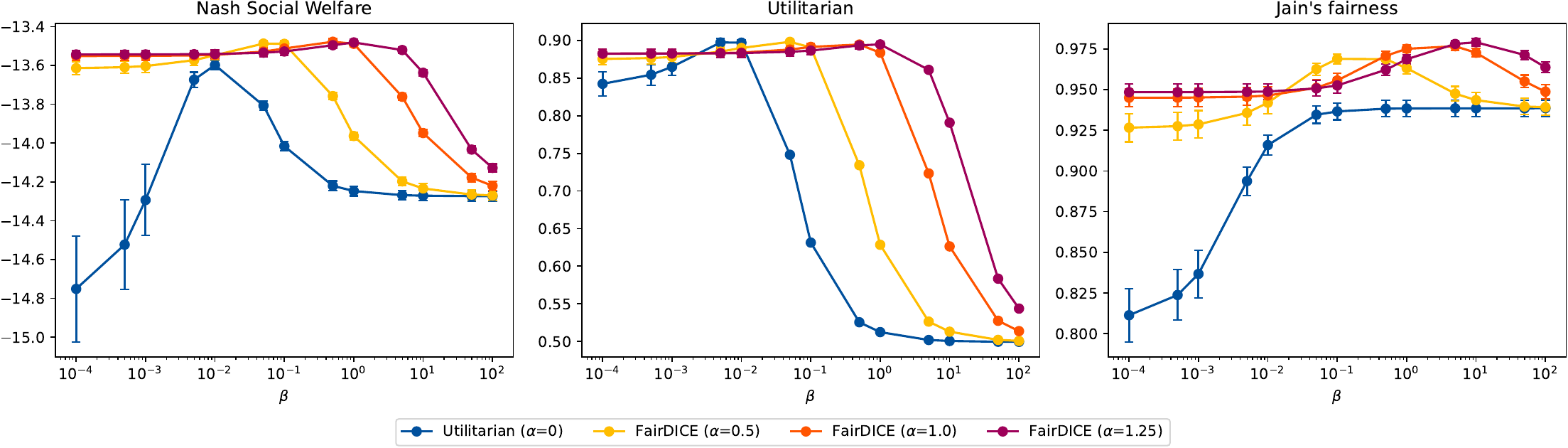}
  \caption{Policy performance on Random MOMDP domain across different $\alpha$ and $\beta$ values, evaluated on Nash social welfare, Utilitarian welfare, and Jain’s fairness index. Results are averaged over 1000 seeds, and reported with 95 $\%$ confidence intervals.}
  \label{fig2}
  \vspace{-0.7cm}
\end{figure}

Figure~\ref{fig2} illustrates how the three performance metrics vary under these trade-offs. Increasing $\alpha$ shifts the objective toward max-min fairness, improving Jain’s fairness index, while decreasing $\alpha$ prioritizes total return, enhancing utilitarian performance. Higher $\beta$ values constrain the learned policies to remain closer to the data collection policy across all metrics. In contrast, lower $\beta$ values reduce regularization, allowing each objective to more effectively pursue its own $\alpha$-fairness. However, if $\beta$ is excessively low, the resulting distribution shift can degrade practical performance.

A surprising finding is that while FairDICE achieves the highest utilitarian performance when maximizing utilitarian welfare ($\alpha = 0$), higher $\alpha$ values also sustain strong utilitarian performance across a wide range of $\beta$. This stems from the concavity of the $\alpha$-fairness objective, where returns contribute less to overall welfare as they grow. Consequently, the incentive for pure return maximization is tempered, implicitly regularizing against distributional shift.

\subsection{Welfare Optimization via Implicit preference weight}
\begin{wrapfigure}{r}{0.3\textwidth}
    \vspace{-0.3cm}
    \centering
    \includegraphics[width=0.3\textwidth]{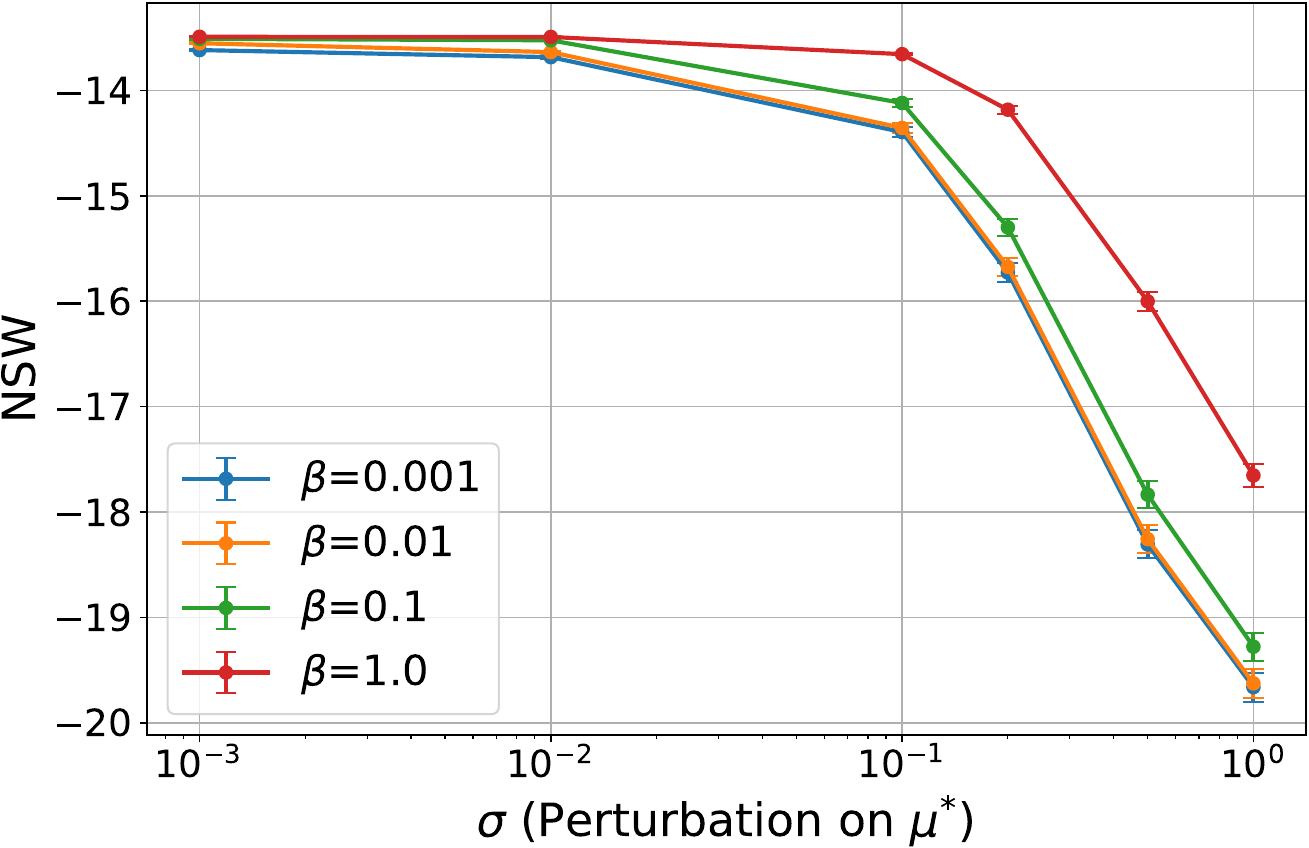}
    \caption{FairDICE-fixed with perturbed $\mu^{*}$}
    \label{fig:nsw_sigma}
    \vspace{-0.3cm}
\end{wrapfigure}

In Section~\ref{sec:eqlinear}, we established the equivalence between regularized linear and fair MORL. We validate this by perturbing the optimal $\mu^{*}$ obtained by FairDICE-NSW in the Random MOMDP experiment and apply it to FairDICE-fixed, as reported in Figure~\ref{fig:nsw_sigma}. Gaussian noise with standard deviation $\sigma$ is generated and applied to each dimension of $\mu^{*}$ by scaling it as $(1 + \text{noise})$. FairDICE-fixed achieves the highest NSW without perturbation, and the NSW decreases as the preference weights deviate from the optimal value. This indicates that our regularized welfare optimization framework shares the same optimal solution with regualrized MORL with linear scalarization and optimizes its implicit preference that maximize NSW.

\begin{figure}[H]
  \vspace{+0.1cm}
  \centering
  \includegraphics[width=0.95\linewidth]{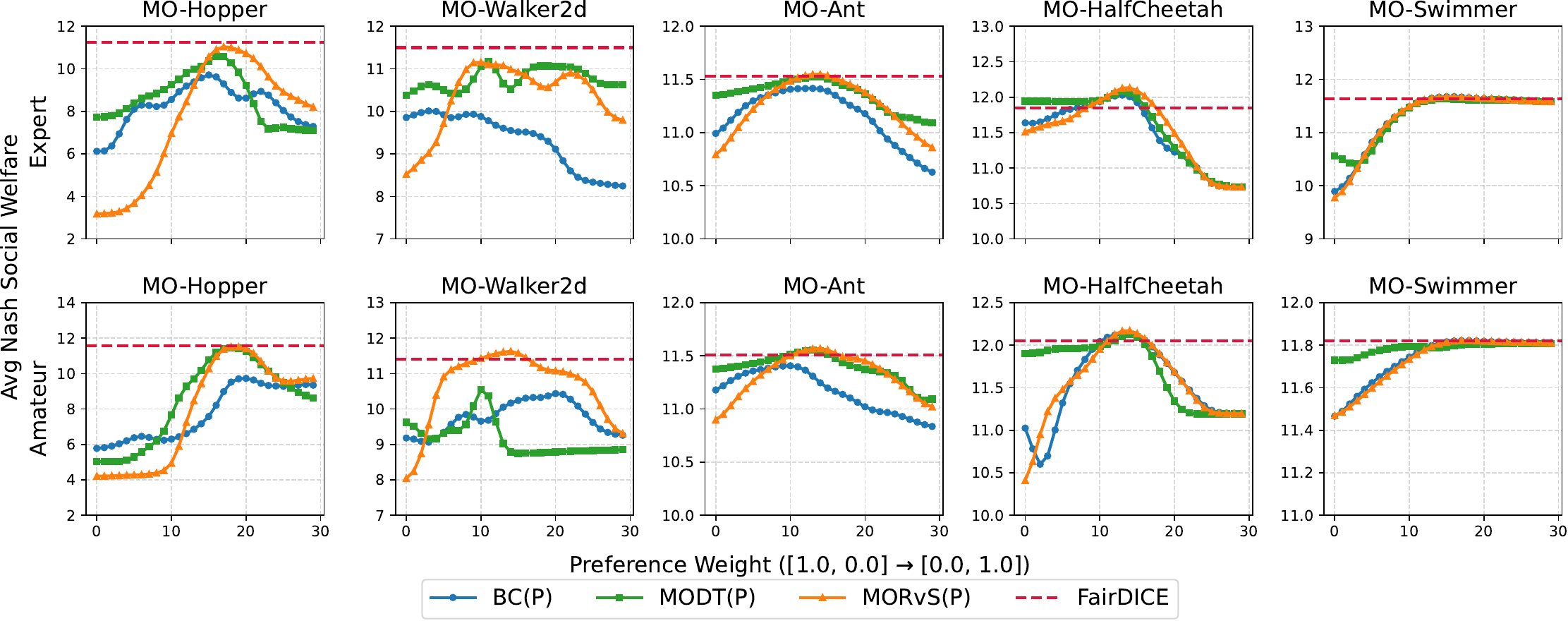}
  \caption{
    Nash social welfare scores for five two-objective tasks, evaluated across 30 linearly spaced preference weights. Each curve shows the average NSW over 5 seeds and 10 evaluation episodes per seed. Red line indicates the average NSW performance of FairDICE.
  }
  \label{fig:nsw_score_dist}
  \vspace{-0.3cm}
\end{figure}

\section{Welfare Maximization in Continuous Domains}
\label{experiments}

\paragraph{Environments} 
We evaluate our method on the D4MORL benchmark \cite{zhu2023scaling}, a standard MORL benchmark in continuous control domains. D4MORL builds upon the D4RL benchmark \cite{fu2020d4rl} by decomposing the original MuJoCo rewards into multiple objectives, such as speed, height and energy efficiency. The dataset for each domain consists of two types of data, collected using either expert or stochastically perturbed (amateur) behavioral policies, and is annotated with preference vectors. Our main experiments include five two-objective tasks (e.g., MO-Hopper, MO-Walker2d, MO-HalfCheetah, MO-Ant, MO-Swimmer) and one three-objective task (MO-Hopper-3obj). Further details on the environments and experiments are  reported in Appendix~\ref{appendix:d4morl}.

\paragraph{Baselines}
While FairDICE seeks a single offline RL policy that maximizes welfare, no existing algorithm directly optimizes this objective in offline MORL. Therefore, we compare against three offline MORL approaches with linear scalarization, additionally searching for preference weights that maximize NSW by uniformly discretizing the simplex and evaluating NSW at each point. Specifically, we adopt three baselines that learn preference-conditioned policies \cite{zhu2023scaling}. 
\begin{itemize}
    \item \textbf{BC(P)} performs behavioral cloning by conditioning on the state and a preference vector to imitate observed actions.
    \item \textbf{MODT(P)} extends Decision Transformer by modeling trajectories as sequences of (state, action, return-to-go) tokens concatenated with a preference vector, using a transformer to predict actions.
    \item \textbf{MORvS(P)} simplifies this setup using a feedforward model that takes the current state and preference-weighted return-to-go as input, enabling more efficient training.
\end{itemize}
While the baseline includes approaches that do not concatenate the state and linear preference ratio, we do not evaluate them as they generally underperform compared to their preference-conditioned counterparts. In contrast, our method does not assume behavior policy preference weights but directly optimizes them to maximize welfare, making it applicable even to datasets without this information.

\begin{figure}
  \centering
  \includegraphics[width=\linewidth]{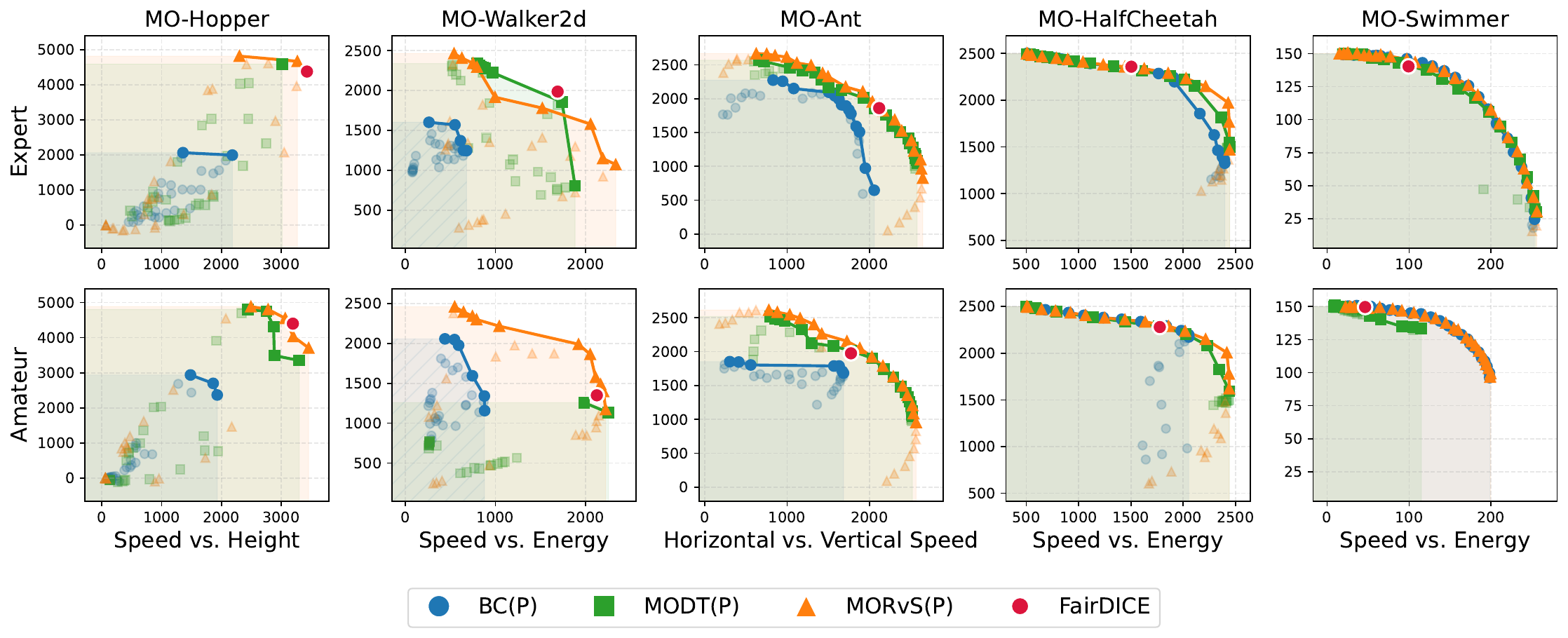}
  \caption{
    Raw return evaluations on five two-objective MuJoCo tasks from D4MORL. Each point represents policy performance under a specific preference weight; Pareto frontiers and dominated regions are shown.
  }
  \label{fig:scatter-continuous}
    \vspace{-0.2cm}
\end{figure}

\paragraph{Offline Fair MORL Performance}
Figure~\ref{fig:nsw_score_dist} shows that FairDICE achieves competitive or superior NSW performance across all two-objective D4MORL tasks, compared to the best results obtained by extensively searching over preference weights in existing methods. Since NSW summarizes multiple objectives into a single scalar, it does not fully capture how well each objective is optimized. To better illustrate FairDICE’s effectiveness across all objectives, we also position its raw returns relative to the Pareto frontier formed by preference-conditioned baselines. Figure~\ref{fig:scatter-continuous} shows that the FairDICE solution lies on the Pareto frontier, highlighting the strong practical performance of our approach.

\begin{figure}
  \centering
  \includegraphics[width=\linewidth]{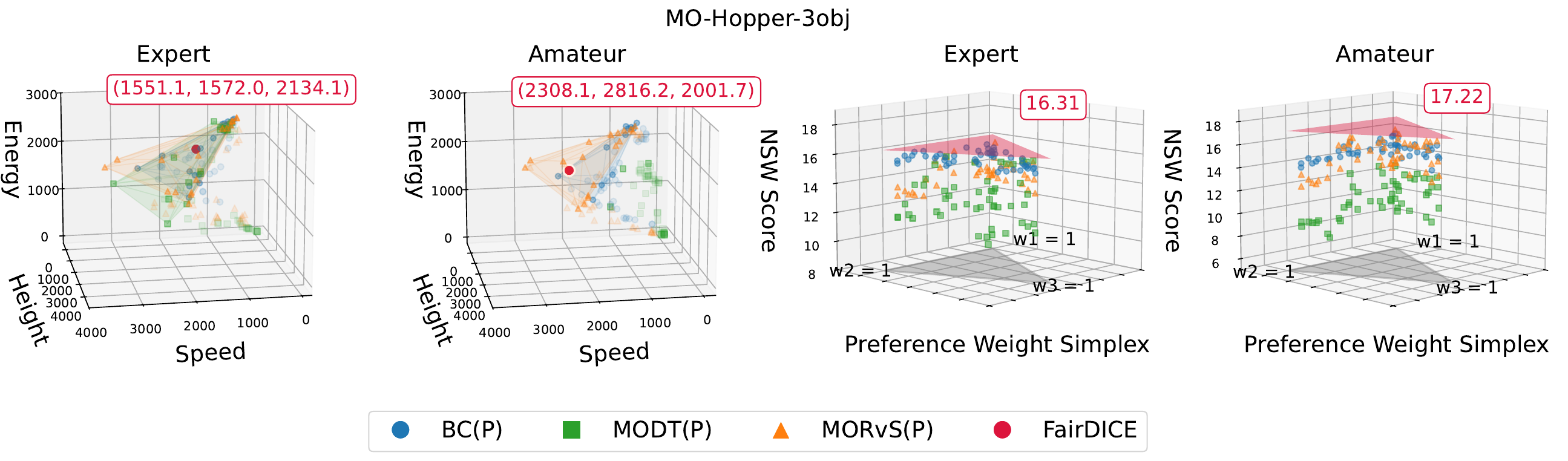}
  \caption{
    Raw returns and Nash social welfare evaluations on MO-Hopper-3obj with three objectives: speed, height, and energy. 50 preference weights are sampled uniformly from the 3D simplex. Red plane indicates average NSW performance of FairDICE.
  }
  \label{fig:pareto_nsw_v3}
  \vspace{-0.4cm}
\end{figure}

In the MO-Hopper-3obj task with three objectives, the preference weight space expands substantially, making it increasingly difficult to find weights that maximize NSW using MORL with linear scalarization. However, Figure~\ref{fig:pareto_nsw_v3} illustrates that FairDICE identifies a welfare-maximizing policy without requiring explicit preference conditioning. Notably, while optimizing for NSW, the resulting policy also achieves raw returns that lie close to or even surpass the Pareto frontier formed by preference-conditioned baselines, indicating that FairDICE achieves strong efficiency in addition to fairness. Moreover, a single additional scalar parameter is sufficient to handle the increased number of objectives, highlighting the scalability and practical efficiency of FairDICE, particularly in high-dimensional preference spaces.

\section{Conclusion}
\label{conclusion}
In this paper, we introduce a novel regularized welfare optimization framework for maximizing welfare in offline MORL, enabling fair outcomes across objectives using fixed datasets—a setting not addressed by prior work. We establish a theoretical connection between regularized MORL with linear scalarization, showing that our framework implicitly learns preference weights that maximize welfare. Building on this, we extend the DICE RL framework to derive our sample based algorithm, \textbf{FairDICE}, that overcomes the optimization challenge caused by the nonlinearity of the objective. Empirically, FairDICE achieves strong fairness aware performance across both discrete and continuous domains with a fixed dataset, effectively balancing trade-offs between objectives and between welfare and distributional shift.

\paragraph{Limitation} 
While our method effectively handles MORL with strictly concave scalarization functions, it does not cover all forms of nonlinear scalarization. Our formulation assumes convexity, and thus the DICE-based approach relying on the Lagrangian dual does not apply to non-convex scalarization in MORL. Additionally, although concave scalarization helps mitigate sensitivity to distribution shift, as an offline RL method, the final performance still depends on the choice of hyperparameters controlling distribution shift and the quality of the dataset.

\section{Acknowledgements}
This work was partly supported by Institute of Information $\&$ Communications Technology Planning $\&$ Evaluation (IITP) grant funded by the Korea government (MSIT) (No. RS-2022-II220311, Development of Goal-Oriented Reinforcement Learning Techniques for Contact-Rich Robotic Manipulation of Everyday Objects, No. RS-2024-00457882, AI Research Hub Project, No. RS-2019-II190079, Artificial Intelligence Graduate School Program (Korea University), and No. RS-2025-25410841, Beyond the Turing Test: Human-Level Game-Playing Agents with Generalization and Adaptation),
the IITP (Institute of Information $\&$ Communications Technology Planning $\&$ Evaluation)-ITRC (Information Technology Research Center) grant funded by the Korea government (Ministry of Science and ICT) (IITP-2025-RS-2024-00436857),
the NRF (RS-2024-00451162) funded by the Ministry of Science and ICT, Korea,
BK21 Four project of the National Research Foundation of Korea, 
the National Research Foundation of Korea (NRF) grant funded by the Korea government (MSIT) (RS-2025-00560367, RS-2025-24803384),
the IITP under the Artificial Intelligence Star Fellowship support program to nurture the best talents (IITP-2025-RS-2025-02304828) grant funded by the Korea government (MSIT),
and KOREA HYDRO $\&$ NUCLEAR POWER CO., LTD (No. 2024-Tech-09).
This work was also supported by the Institute of Information $\&$ Communications Technology Planning $\&$ Evaluation (IITP) grant (RS-2020-II201361, Artificial Intelligence Graduate School Program (Yonsei University) and the Yonsei University Research Fund of 2025-22-0158.



\bibliography{neurips_2025}

\section*{NeurIPS Paper Checklist}

\begin{enumerate}

\item {\bf Claims}
    \item[] Question: Do the main claims made in the abstract and introduction accurately reflect the paper's contributions and scope?
    \item[] Answer: \answerYes{} 
    \item[] Justification: Section \ref{intro}
    \item[] Guidelines:
    \begin{itemize}
        \item The answer NA means that the abstract and introduction do not include the claims made in the paper.
        \item The abstract and/or introduction should clearly state the claims made, including the contributions made in the paper and important assumptions and limitations. A No or NA answer to this question will not be perceived well by the reviewers. 
        \item The claims made should match theoretical and experimental results, and reflect how much the results can be expected to generalize to other settings. 
        \item It is fine to include aspirational goals as motivation as long as it is clear that these goals are not attained by the paper. 
    \end{itemize}

\item {\bf Limitations}
    \item[] Question: Does the paper discuss the limitations of the work performed by the authors?
    \item[] Answer: \answerYes{} 
    \item[] Justification: Section \ref{conclusion}
    \item[] Guidelines:
    \begin{itemize}
        \item The answer NA means that the paper has no limitation while the answer No means that the paper has limitations, but those are not discussed in the paper. 
        \item The authors are encouraged to create a separate "Limitations" section in their paper.
        \item The paper should point out any strong assumptions and how robust the results are to violations of these assumptions (e.g., independence assumptions, noiseless settings, model well-specification, asymptotic approximations only holding locally). The authors should reflect on how these assumptions might be violated in practice and what the implications would be.
        \item The authors should reflect on the scope of the claims made, e.g., if the approach was only tested on a few datasets or with a few runs. In general, empirical results often depend on implicit assumptions, which should be articulated.
        \item The authors should reflect on the factors that influence the performance of the approach. For example, a facial recognition algorithm may perform poorly when image resolution is low or images are taken in low lighting. Or a speech-to-text system might not be used reliably to provide closed captions for online lectures because it fails to handle technical jargon.
        \item The authors should discuss the computational efficiency of the proposed algorithms and how they scale with dataset size.
        \item If applicable, the authors should discuss possible limitations of their approach to address problems of privacy and fairness.
        \item While the authors might fear that complete honesty about limitations might be used by reviewers as grounds for rejection, a worse outcome might be that reviewers discover limitations that aren't acknowledged in the paper. The authors should use their best judgment and recognize that individual actions in favor of transparency play an important role in developing norms that preserve the integrity of the community. Reviewers will be specifically instructed to not penalize honesty concerning limitations.
    \end{itemize}

\item {\bf Theory assumptions and proofs}
    \item[] Question: For each theoretical result, does the paper provide the full set of assumptions and a complete (and correct) proof?
    \item[] Answer: \answerYes{} 
    \item[] Justification: Section \ref{framework}
    \item[] Guidelines:
    \begin{itemize}
        \item The answer NA means that the paper does not include theoretical results. 
        \item All the theorems, formulas, and proofs in the paper should be numbered and cross-referenced.
        \item All assumptions should be clearly stated or referenced in the statement of any theorems.
        \item The proofs can either appear in the main paper or the supplemental material, but if they appear in the supplemental material, the authors are encouraged to provide a short proof sketch to provide intuition. 
        \item Inversely, any informal proof provided in the core of the paper should be complemented by formal proofs provided in appendix or supplemental material.
        \item Theorems and Lemmas that the proof relies upon should be properly referenced. 
    \end{itemize}

    \item {\bf Experimental result reproducibility}
    \item[] Question: Does the paper fully disclose all the information needed to reproduce the main experimental results of the paper to the extent that it affects the main claims and/or conclusions of the paper (regardless of whether the code and data are provided or not)?
    \item[] Answer: \answerYes{} 
    \item[] Justification: Section \ref{experiments}, Appendix~\ref{Appendix:Finite}, Appendix ~\ref{appendix:d4morl}, Appendix~\ref{appendix:implementation}
    \item[] Guidelines:
    \begin{itemize}
        \item The answer NA means that the paper does not include experiments.
        \item If the paper includes experiments, a No answer to this question will not be perceived well by the reviewers: Making the paper reproducible is important, regardless of whether the code and data are provided or not.
        \item If the contribution is a dataset and/or model, the authors should describe the steps taken to make their results reproducible or verifiable. 
        \item Depending on the contribution, reproducibility can be accomplished in various ways. For example, if the contribution is a novel architecture, describing the architecture fully might suffice, or if the contribution is a specific model and empirical evaluation, it may be necessary to either make it possible for others to replicate the model with the same dataset, or provide access to the model. In general. releasing code and data is often one good way to accomplish this, but reproducibility can also be provided via detailed instructions for how to replicate the results, access to a hosted model (e.g., in the case of a large language model), releasing of a model checkpoint, or other means that are appropriate to the research performed.
        \item While NeurIPS does not require releasing code, the conference does require all submissions to provide some reasonable avenue for reproducibility, which may depend on the nature of the contribution. For example
        \begin{enumerate}
            \item If the contribution is primarily a new algorithm, the paper should make it clear how to reproduce that algorithm.
            \item If the contribution is primarily a new model architecture, the paper should describe the architecture clearly and fully.
            \item If the contribution is a new model (e.g., a large language model), then there should either be a way to access this model for reproducing the results or a way to reproduce the model (e.g., with an open-source dataset or instructions for how to construct the dataset).
            \item We recognize that reproducibility may be tricky in some cases, in which case authors are welcome to describe the particular way they provide for reproducibility. In the case of closed-source models, it may be that access to the model is limited in some way (e.g., to registered users), but it should be possible for other researchers to have some path to reproducing or verifying the results.
        \end{enumerate}
    \end{itemize}

\item {\bf Open access to data and code}
    \item[] Question: Does the paper provide open access to the data and code, with sufficient instructions to faithfully reproduce the main experimental results, as described in supplemental material?
    \item[] Answer: \answerYes{} 
    \item[] Justification: Appendix~\ref{Appendix:Finite}, Appendix ~\ref{appendix:d4morl}, Appendix~\ref{appendix:implementation}
    \item[] Guidelines:
    \begin{itemize}
        \item The answer NA means that paper does not include experiments requiring code.
        \item Please see the NeurIPS code and data submission guidelines (\url{https://nips.cc/public/guides/CodeSubmissionPolicy}) for more details.
        \item While we encourage the release of code and data, we understand that this might not be possible, so “No” is an acceptable answer. Papers cannot be rejected simply for not including code, unless this is central to the contribution (e.g., for a new open-source benchmark).
        \item The instructions should contain the exact command and environment needed to run to reproduce the results. See the NeurIPS code and data submission guidelines (\url{https://nips.cc/public/guides/CodeSubmissionPolicy}) for more details.
        \item The authors should provide instructions on data access and preparation, including how to access the raw data, preprocessed data, intermediate data, and generated data, etc.
        \item The authors should provide scripts to reproduce all experimental results for the new proposed method and baselines. If only a subset of experiments are reproducible, they should state which ones are omitted from the script and why.
        \item At submission time, to preserve anonymity, the authors should release anonymized versions (if applicable).
        \item Providing as much information as possible in supplemental material (appended to the paper) is recommended, but including URLs to data and code is permitted.
    \end{itemize}

\item {\bf Experimental setting/details}
    \item[] Question: Does the paper specify all the training and test details (e.g., data splits, hyperparameters, how they were chosen, type of optimizer, etc.) necessary to understand the results?
    \item[] Answer: \answerYes{} 
    \item[] Justification: Appendix~\ref{Appendix:Finite}, Appendix ~\ref{appendix:d4morl}, Appendix~\ref{appendix:implementation}
    \item[] Guidelines:
    \begin{itemize}
        \item The answer NA means that the paper does not include experiments.
        \item The experimental setting should be presented in the core of the paper to a level of detail that is necessary to appreciate the results and make sense of them.
        \item The full details can be provided either with the code, in appendix, or as supplemental material.
    \end{itemize}

\item {\bf Experiment statistical significance}
    \item[] Question: Does the paper report error bars suitably and correctly defined or other appropriate information about the statistical significance of the experiments?
    \item[] Answer: \answerYes{} 
    \item[] Justification: Section \ref{empirical}, Appendix~\ref{appendix:robustness}, Appendix~\ref{appendix:alphaablation}
    \item[] Guidelines:
    \begin{itemize}
        \item The answer NA means that the paper does not include experiments.
        \item The authors should answer "Yes" if the results are accompanied by error bars, confidence intervals, or statistical significance tests, at least for the experiments that support the main claims of the paper.
        \item The factors of variability that the error bars are capturing should be clearly stated (for example, train/test split, initialization, random drawing of some parameter, or overall run with given experimental conditions).
        \item The method for calculating the error bars should be explained (closed form formula, call to a library function, bootstrap, etc.)
        \item The assumptions made should be given (e.g., Normally distributed errors).
        \item It should be clear whether the error bar is the standard deviation or the standard error of the mean.
        \item It is OK to report 1-sigma error bars, but one should state it. The authors should preferably report a 2-sigma error bar than state that they have a 96\% CI, if the hypothesis of Normality of errors is not verified.
        \item For asymmetric distributions, the authors should be careful not to show in tables or figures symmetric error bars that would yield results that are out of range (e.g. negative error rates).
        \item If error bars are reported in tables or plots, The authors should explain in the text how they were calculated and reference the corresponding figures or tables in the text.
    \end{itemize}

\item {\bf Experiments compute resources}
    \item[] Question: For each experiment, does the paper provide sufficient information on the computer resources (type of compute workers, memory, time of execution) needed to reproduce the experiments?
    \item[] Answer: \answerYes{}{} 
    \item[] Justification: Appendix~\ref{appendix:Resources}
    \item[] Guidelines:
    \begin{itemize}
        \item The answer NA means that the paper does not include experiments.
        \item The paper should indicate the type of compute workers CPU or GPU, internal cluster, or cloud provider, including relevant memory and storage.
        \item The paper should provide the amount of compute required for each of the individual experimental runs as well as estimate the total compute. 
        \item The paper should disclose whether the full research project required more compute than the experiments reported in the paper (e.g., preliminary or failed experiments that didn't make it into the paper). 
    \end{itemize}
    
\item {\bf Code of ethics}
    \item[] Question: Does the research conducted in the paper conform, in every respect, with the NeurIPS Code of Ethics \url{https://neurips.cc/public/EthicsGuidelines}?
    \item[] Answer: \answerYes{} 
    \item[] Justification: We have reviewed the NeurIPS Code for Ethics
    \item[] Guidelines:
    \begin{itemize}
        \item The answer NA means that the authors have not reviewed the NeurIPS Code of Ethics.
        \item If the authors answer No, they should explain the special circumstances that require a deviation from the Code of Ethics.
        \item The authors should make sure to preserve anonymity (e.g., if there is a special consideration due to laws or regulations in their jurisdiction).
    \end{itemize}

\item {\bf Broader impacts}
    \item[] Question: Does the paper discuss both potential positive societal impacts and negative societal impacts of the work performed?
    \item[] Answer: \answerNA{} 
    \item[] Justification: This paper does not directly engage with societal elements or have immediate applications that could be identified as having positive or negative societal impacts.
    \item[] Guidelines:
    \begin{itemize}
        \item The answer NA means that there is no societal impact of the work performed.
        \item If the authors answer NA or No, they should explain why their work has no societal impact or why the paper does not address societal impact.
        \item Examples of negative societal impacts include potential malicious or unintended uses (e.g., disinformation, generating fake profiles, surveillance), fairness considerations (e.g., deployment of technologies that could make decisions that unfairly impact specific groups), privacy considerations, and security considerations.
        \item The conference expects that many papers will be foundational research and not tied to particular applications, let alone deployments. However, if there is a direct path to any negative applications, the authors should point it out. For example, it is legitimate to point out that an improvement in the quality of generative models could be used to generate deepfakes for disinformation. On the other hand, it is not needed to point out that a generic algorithm for optimizing neural networks could enable people to train models that generate Deepfakes faster.
        \item The authors should consider possible harms that could arise when the technology is being used as intended and functioning correctly, harms that could arise when the technology is being used as intended but gives incorrect results, and harms following from (intentional or unintentional) misuse of the technology.
        \item If there are negative societal impacts, the authors could also discuss possible mitigation strategies (e.g., gated release of models, providing defenses in addition to attacks, mechanisms for monitoring misuse, mechanisms to monitor how a system learns from feedback over time, improving the efficiency and accessibility of ML).
    \end{itemize}
    
\item {\bf Safeguards}
    \item[] Question: Does the paper describe safeguards that have been put in place for responsible release of data or models that have a high risk for misuse (e.g., pretrained language models, image generators, or scraped datasets)?
    \item[] Answer: \answerNA{} 
    \item[] Justification: This paper poses no such risks.
    \item[] Guidelines:
    \begin{itemize}
        \item The answer NA means that the paper poses no such risks.
        \item Released models that have a high risk for misuse or dual-use should be released with necessary safeguards to allow for controlled use of the model, for example by requiring that users adhere to usage guidelines or restrictions to access the model or implementing safety filters. 
        \item Datasets that have been scraped from the Internet could pose safety risks. The authors should describe how they avoided releasing unsafe images.
        \item We recognize that providing effective safeguards is challenging, and many papers do not require this, but we encourage authors to take this into account and make a best faith effort.
    \end{itemize}

\item {\bf Licenses for existing assets}
    \item[] Question: Are the creators or original owners of assets (e.g., code, data, models), used in the paper, properly credited and are the license and terms of use explicitly mentioned and properly respected?
    \item[] Answer: \answerYes{} 
    \item[] Justification:  Appendix ~\ref{appendix:d4morl}, Appendix~\ref{appendix:implementation}
    \item[] Guidelines:
    \begin{itemize}
        \item The answer NA means that the paper does not use existing assets.
        \item The authors should cite the original paper that produced the code package or dataset.
        \item The authors should state which version of the asset is used and, if possible, include a URL.
        \item The name of the license (e.g., CC-BY 4.0) should be included for each asset.
        \item For scraped data from a particular source (e.g., website), the copyright and terms of service of that source should be provided.
        \item If assets are released, the license, copyright information, and terms of use in the package should be provided. For popular datasets, \url{paperswithcode.com/datasets} has curated licenses for some datasets. Their licensing guide can help determine the license of a dataset.
        \item For existing datasets that are re-packaged, both the original license and the license of the derived asset (if it has changed) should be provided.
        \item If this information is not available online, the authors are encouraged to reach out to the asset's creators.
    \end{itemize}

\item {\bf New assets}
    \item[] Question: Are new assets introduced in the paper well documented and is the documentation provided alongside the assets?
    \item[] Answer: \answerNA{} 
    \item[] Justification: This paper does not release new assets.
    \item[] Guidelines:
    \begin{itemize}
        \item The answer NA means that the paper does not release new assets.
        \item Researchers should communicate the details of the dataset/code/model as part of their submissions via structured templates. This includes details about training, license, limitations, etc. 
        \item The paper should discuss whether and how consent was obtained from people whose asset is used.
        \item At submission time, remember to anonymize your assets (if applicable). You can either create an anonymized URL or include an anonymized zip file.
    \end{itemize}

\item {\bf Crowdsourcing and research with human subjects}
    \item[] Question: For crowdsourcing experiments and research with human subjects, does the paper include the full text of instructions given to participants and screenshots, if applicable, as well as details about compensation (if any)? 
    \item[] Answer: \answerNA{} 
    \item[] Justification: This paper does not involve crowdsourcing nor research with human subjects.
    \item[] Guidelines:
    \begin{itemize}
        \item The answer NA means that the paper does not involve crowdsourcing nor research with human subjects.
        \item Including this information in the supplemental material is fine, but if the main contribution of the paper involves human subjects, then as much detail as possible should be included in the main paper. 
        \item According to the NeurIPS Code of Ethics, workers involved in data collection, curation, or other labor should be paid at least the minimum wage in the country of the data collector. 
    \end{itemize}

\item {\bf Institutional review board (IRB) approvals or equivalent for research with human subjects}
    \item[] Question: Does the paper describe potential risks incurred by study participants, whether such risks were disclosed to the subjects, and whether Institutional Review Board (IRB) approvals (or an equivalent approval/review based on the requirements of your country or institution) were obtained?
    \item[] Answer: \answerNA{} 
    \item[] Justification: This paper does not involve crowdsourcing nor research with human subjects.
    \item[] Guidelines:
    \begin{itemize}
        \item The answer NA means that the paper does not involve crowdsourcing nor research with human subjects.
        \item Depending on the country in which research is conducted, IRB approval (or equivalent) may be required for any human subjects research. If you obtained IRB approval, you should clearly state this in the paper. 
        \item We recognize that the procedures for this may vary significantly between institutions and locations, and we expect authors to adhere to the NeurIPS Code of Ethics and the guidelines for their institution. 
        \item For initial submissions, do not include any information that would break anonymity (if applicable), such as the institution conducting the review.
    \end{itemize}

\item {\bf Declaration of LLM usage}
    \item[] Question: Does the paper describe the usage of LLMs if it is an important, original, or non-standard component of the core methods in this research? Note that if the LLM is used only for writing, editing, or formatting purposes and does not impact the core methodology, scientific rigorousness, or originality of the research, declaration is not required.
    \item[] Answer: \answerNA{} 
    \item[] Justification: The core method development in this research does not involve LLMs as any important, original, or non-standard components.
    \item[] Guidelines:
    \begin{itemize}
        \item The answer NA means that the core method development in this research does not involve LLMs as any important, original, or non-standard components.
        \item Please refer to our LLM policy (\url{https://neurips.cc/Conferences/2025/LLM}) for what should or should not be described.
    \end{itemize}

\end{enumerate}

\newpage
\appendix
\section{DICE-RL framework}
\label{Appendix:dicerlframework}
In this section, we introduce the DICE-RL framework along with its corresponding offline single-objective reinforcement learning algorithm, OptiDICE, as proposed in \citep{lee2021optidice}. Our algorithm can be viewed as a multi-objective extension of OptiDICE. The DICE-RL framework is an offline RL framework where return maximization is regularized with an $f$-divergence between the stationary distribution of the learned policy $d$ and the empirical distribution $d_D$, solving:
 \begin{align}
\max_{d \geq 0} \ & \sum_{s,a} d(s,a) r(s,a) - \beta \sum_{s,a} d_D(s,a) f\left(\frac{d(s,a)}{d_D(s,a)}\right)\nonumber \\
\text{s.t.} \ & \sum_{a}d(s,a) = (1-\gamma)p_0(s) + \gamma\sum_{\bar{s},\bar{a}}T(s|\bar{s},\bar{a})d(\bar{s},\bar{a}), \quad \forall s \nonumber
\end{align}
where the Bellman flow constraints ensure that the optimal $d^{*}(s,a)$ constitutes a valid stationary distribution. Lagrangian dual of the convex optimization problem is given by,
\begin{align}
\max_{d\geq 0}\min_{\nu} \ &\sum_{s,a}d(s,a)r(s,a) + \sum_{s}\nu(s)\mathcal{F}_{d}(s)- \beta \sum_{s,a} d_D(s,a) f\left(\frac{d(s,a)}{d_D(s,a)}\right)\nonumber
\end{align}
where $\mathcal{F}_d(s)= (1-\gamma)p_0(s)+\gamma\sum_{\bar{s},\bar{a}}T(s|\bar{s},\bar{a})d(\bar{s},\bar{a})-\sum_{a}d(s,a)$.
We reparameterize the stationary distribution as $d(s,a)=w(s,a)d_D(s,a)$ and express the dual in terms of the importance weights $w$, using the identity $\sum_{s}T(s|\bar{s},\bar{a})\nu(s)\sum_{\bar{s},\bar{a}}d(\bar{s},\bar{a}) = \sum_{s,a}d(s,a)\sum_{s'}T(s'|s,a)\nu(s')$. This yields the following optimization problem:
\begin{align}
&\max_{w\geq 0}\min_{\nu} \ L(w,\nu) := \mathbb{E}_{s\sim p_0}[(1-\gamma)\nu(s)]+\mathbb{E}_{(s,a)\sim d_D}\left[w(s,a)e_{\nu}(s,a)-\beta f\left(w(s,a)\right)\right]\nonumber
\end{align}
where $e_{\nu}(s,a)=r(s,a)+\gamma\sum_{s'}T(s'|s,a)\nu(s')-\nu(s) \ \forall s,a$.

Strong duality holds by Slater’s condition, since the problem is convex and a valid stationary distribution exists in the strictly feasible set. Therefore, the order of optimization can be swapped to $\min_{\nu}\max_{w\geq0}$. Optimal $w^{*}(s,a)$ is computed from the first-order condition given by:
\begin{align}
    \frac{\partial L(w,\nu)}{\partial w(s,a)}=\mathbb{E}_{(s,a)\sim d_D}\left[e_{\nu}(s,a)-\beta f'\left(w(s,a)\right)\right]=0\nonumber
\end{align}
where $w^{*}_{\nu}(s,a) = \max \left(0, (f')^{-1}\left(\frac{e_{\nu}(s,a)}{\beta}\right)\right)$. By plugging $w^{*}(s,a)$ in $L(w,\nu)$ results in the following $\nu$ loss of OptiDICE.
\begin{align}
\min_{\nu} \ \mathbb{E}_{s\sim p_0}[(1-\gamma)\nu(s)]+\mathbb{E}_{(s,a)\sim d_D}\left[\beta f^{*}_{0}\left(\frac{e_{\nu}(s,a)}{\beta}\right)\right] \nonumber\end{align} 
where $f^{*}_{0}(y):=\max_{x\geq0}xy-f(x)$.

After minimizing over $\nu$, the optimal stationary distribution is given by $d^{*}(s,a) = w^{*}_{\nu^{*}}(s,a)d_D(s,a)$. This distribution can then be used to recover the optimal policy that induces $d^{*}(s,a)$ via $\pi^{*}(a|s) = \frac{d^{*}(s,a)}{\sum_{a} d^{*}(s,a)}$ for all $s,a$, or through a policy extraction method such as weighted behavior cloning.

\section{Counterexample}
\label{Appendix:counter}
In this section, we present a counterexample demonstrating that applying the optimal implicit weights $\mu^{*}$ from Fair MORL (P2) to Linear MORL (P3) does not recover the optimal policy of Fair MORL (P2). Consider an MDP with a single state $s$ and a terminal state reached immediately after taking an action. There are two available actions, $a \in \{a_1, a_2\}$, with corresponding stationary distributions $d(s, a_1)$ and $d(s, a_2)$. Each action yields a reward vector consisting of rewards for objectives $A$ and $B$: $\mathbf{r}(s,a_1)=[r_A(s,a_1), r_B(s,a_1)]=[1,4]$ and $\mathbf{r}(s,a_2)=[r_A(s,a_2), r_B(s,a_2)]=[3,1]$.

Using this setup, we construct the corresponding (P1) optimization problem, while assuming Nash social welfare ($u_{i}= \log(x) \ \forall i$):
\begin{align}
    \text{(P1): } \max_{d\geq0}\;&\log \left(d(s,a_{1})+ 3d(s,a_{2})\right)+ \log \left(4d(s,a_{1})+ d(s,a_{2})\right)\nonumber\\
\text{s.t.}\;&d(s,a_{1})+d(s,a_{2})=1\nonumber
\end{align}
where $\log(x)$ is applied to the returns of each objective. An equivalent optimization (P2) is given by:
\begin{align}
    \text{(P2): } \max_{d\geq0,k}\;&\log \left(k_1\right)+ \log \left(k_2\right)\nonumber\\
    \text{s.t.}\;&d(s,a_{1})+d(s,a_{2})=1\nonumber\\
    &d(s,a_{1})+ 3d(s,a_{2}) = k_1 \nonumber\\
    &4d(s,a_{1})+ d(s,a_{2}) = k_2 \nonumber
\end{align}
Its Lagrangian dual is given as,
\begin{align}
\max_{d\geq0,k}\min_{\mu,\nu} \ &\log \left(k_1\right)+ \log \left(k_2)\right) +\nu(d(s,a_{1})+d(s,a_{2})-1) \nonumber\\ &+\mu_{1}(d(s,a_{1})+ 3d(s,a_{2})-k_1) +\mu_{2} (4d(s,a_{1})+ d(s,a_{2}) - k_2)\nonumber
\end{align}
By solving the first-order conditions, the optimal Lagrange multipliers are $\mu_{1}^{*} \approx 0.5455$ and $\mu_{2}^{*} \approx 0.3636$, resulting in the optimal stationary distribution $[d^{*}(s, a_1), d^{*}(s, a_2)] = [0.5834, 0.4166]$. The optimal policy of Fair MORL (P2) is a stochastic policy that prefers action $a_{1}$ while still assigning probability to $a_{2}$, resulting in objective returns $k^{*} = [1.8332, 2.7500]$. While $\mu$ is applied to the objective returns in a manner analogous to Linear MORL, we demonstrate that the two formulations are fundamentally distinct by applying the implicit preference weight $\mu^{*}$ to (P3):
\begin{align}
    \text{(P3): } \max_{d\geq0}\;&(1\cdot\mu^{*}_{1}+4\cdot\mu^{*}_2)d(s,a_{1})+ (3\cdot\mu^{*}_{1}+1\cdot\mu^{*}_2)d(s,a_{2})\nonumber\\
\text{s.t.}\;&d(s,a_{1})+d(s,a_{2})=1\nonumber
\end{align}

In this case, $1\cdot\mu^{*}_{1}+4\cdot\mu^{*}_2=3\cdot\mu^{*}_{1}+1\cdot\mu^{*}_2=2.0$, indicating that all policies are the all optimal policies of (P3). As a result, any policy is optimal under Linear MORL with fixed weights $\mu^{}$. This demonstrates that (P2) and (P3) are distinct optimization problems.

\section{Proof of Proposition~\ref{prop1}}
\label{Appendix:propositionproof}
In this section, we provide a proof of Proposition~\ref{prop1}, showing that regularized Linear MORL (P3-reg), when using preference weights equal to the optimal dual variable $\mu^{*}$ from regularized Fair MORL (P2-reg), converges to the same solution as (P2-reg). To facilitate the explanation, we begin by rewriting the (P3-reg) formulation:
\begin{align}
    \text{(P3-reg): } \max_{d\geq0}\;&\sum_{s,a}d(s,a) \sum_{i}\mu^{*}_{i}r_i(s,a)-\beta\sum_{s,a}d_D(s,a)f\left(\frac{d(s,a)}{d_D(s,a)}\right)\quad
\text{s.t.}\;\eqref{const:Bellman} \nonumber
\end{align}
The Lagrangian duals of (P2-reg) and (P3-reg) are given by:
\begin{align}
\max_{d\geq 0, k}\min_{\nu, \mu} \ &L_{\text{P2-reg}}(\nu,\mu,d,k) := \sum_{i}\mu_{i}\bigg(\sum_{s,a}d(s,a)r_i(s,a)-k_i\bigg) -\beta \sum_{s,a}d_D(s,a) f\left(\frac{d(s,a)}{d_D(s,a)}\right) \nonumber \\
&\quad \quad \quad \quad \quad \quad \quad \quad+ \sum_{i}u_{i}\left(k_i\right) +\sum_s \nu(s)\mathcal{F}_d(s) \nonumber\\
\max_{d\geq 0}\min_{\nu} \ &L_{\text{P3-reg}}(\nu,d) := \sum_{s,a}d(s,a) \sum_{i}\mu^{*}_{i}r_i(s,a)-\beta \sum_{s,a}d_D(s,a) f\left(\frac{d(s,a)}{d_D(s,a)}\right) \nonumber\\ &\quad \quad \quad \quad \quad \quad \quad \quad+\sum_s \nu(s)\mathcal{F}_d(s) \nonumber
\end{align}
Assuming the optimal $\mu^{*}$ from (P2-reg) is given, we compute the gradient of each Lagrangian with respect to $\nu(s)$ and $d(s, a)$, and show that both share the same gradient at their optimal solutions.
\begin{align}
    \frac{\partial L_{\text{P2-reg}}}{\partial \nu(s)} = \frac{\partial L_{\text{P3-reg}}}{\partial \nu(s)} &= \mathcal{F}_{d^{*}}(s)\nonumber \\
    \frac{\partial L_{\text{P2-reg}}}{\partial d(s,a)} = \frac{\partial L_{\text{P3-reg}}}{\partial d(s,a)} &= \sum_{i}\mu^{*}_{i}r_{i}(s,a) - \beta \sum_{s,a}f'\left(\frac{d^{*}(s,a)}{d_D(s,a)}\right) +\sum_{s'}\gamma T(s'|s,a)\nu^{*}(s')-\nu^{*}(s)\nonumber
\end{align}
While (P3) is a linear program, (P3-reg) becomes a convex optimization problem due to regularization with a strictly convex function $f$, making the KKT conditions applicable. From the stationarity conditions, (P3-reg) with $\mu^{*}$ converges to the same optimal solution as (P2-reg).
\subsection{Empirical evidence}
We adapt the counterexample from Appendix~\ref{Appendix:counter} to demonstrate that, under offline regularization, Linear MORL and Fair MORL can share the same optimal solution. In the offline setting, we additionally assume a fixed data distribution over actions given by $[d_D(s,a_1), d_D(s,a_2)] = [0.7, 0.3]$. The (P2-reg) formulation of the counterexample is given by:
\begin{align}
    \text{(P2-reg): } \max_{d\geq0,k}\;&\log \left(k_1\right)+ \log \left(k_2\right)-\sum_{a}d_D(s,a)f\left(\frac{d(s,a)}{d_D(s,a)}\right)\nonumber\\
    \text{s.t.}\;&d(s,a_{1})+d(s,a_{2})=1\nonumber\\
    &d(s,a_{1})+ 3d(s,a_{2}) = k_1 \nonumber\\
    &4d(s,a_{1})+ d(s,a_{2}) = k_2 \nonumber
\end{align}
where we adopt $\chi^2$-divergence $f(x)=\frac{1}{2}(x-1)^2$.
The optimal Lagrange multipliers are $\mu_{1}^{*} \approx 0.5959$ and $\mu_{2}^{*} \approx 0.3352$, resulting in the optimal stationary distribution $[d^{*}(s, a_1), d^{*}(s, a_2)] = [0.6609, 0.3390]$. This indicates that as $a_1$ is more common in the data distribution than $a_2$, offline RL policy of (P2-reg) favors $a_1$ compared to the unregularized case. This aligns with the goal of offline reinforcement learning where the optimized stationary distribution should not deviate excessively from the dataset distribution. We apply $\mu^{*}$ to (P3-reg).
\begin{align}
    \text{(P3-reg): } \max_{d\geq0}\;&(1\cdot\mu^{*}_{1}+4\cdot\mu^{*}_2)d(s,a_{1})+ (3\cdot\mu^{*}_{1}+1\cdot\mu^{*}_2)d(s,a_{2})-\sum_{a}d_D(s,a)f\left(\frac{d(s,a)}{d_D(s,a)}\right)\nonumber\\
\text{s.t.}\;&d(s,a_{1})+d(s,a_{2})=1\nonumber
\end{align}
where $1\cdot\mu^{*}_{1}+4\cdot\mu^{*}_2=1.936$ and $3\cdot\mu^{*}_{1}+1\cdot\mu^{*}_2=2.123$.

As (P3-reg) is a convex optimization problem with strictly concave objective, the uniqueness of the optimal solution is guaranteed. The optimal stationary distribution of (P3-reg) is equivalent to that of (P2-reg) as $[d^{*}(s, a_1), d^{*}(s, a_2)] = [0.6609, 0.3390]$. This establishes the connection between regularized Linear MORL and Fair MORL theoretically and empirically.

\section{Finite Domain Experiment Setting}
\label{Appendix:Finite}
In this section, we provide a detailed explanation of the experimental setup used in Section~\ref{empirical}. We begin by describing how the classic Four-Room environment~\cite{lee2021optidice, sutton1999between} and Random MDP~\cite{lee2021optidice, laroche2019safe} are adapted for the offline multi-objective reinforcement learning (MORL) setting. We present additional visualizations of the MO-Four-Room results to further support the findings in Section~\ref{empirical}.
\subsection{Environment detail}
\paragraph{MO-Four-Room} In MO-Four-Room domain, three distinct goals, each associated with a separate objective. The agent starts from the initial state (orange) and navigates toward one of the goal states (green). Upon reaching a goal, it receives a one-hot reward vector: $[1, 0, 0]$ for the lower-left room, $[0, 1, 0]$ for the upper-right room, and $[0, 0, 1]$ for the lower-right room. The agent selects one of four actions, $\{\text{left}, \text{right}, \text{up}, \text{down}\}$; however, the environment is stochastic, and with a probability of 0.1, the agent transitions in a different direction than the one intended. To simulate offline RL, a dataset of 300 trajectories are collected from a uniformly random behavior policy. The experiments are conducted with $\alpha \in \{0.0, 1.0\}$, where $\alpha = 0.0$ corresponds to a policy that maximizes Utilitarian welfare, and $\alpha = 1.0$ corresponds to one that maximizes Nash social welfare. The regularization coefficient is fixed at $\beta = 0.01$ and $f(x) = 0.5(x-1)^2$.

\paragraph{Random MOMDP} In the Random MOMDP domain, a multi-objective Markov decision process is generated with $|\mathcal{S}| = 50$, $|\mathcal{A}| = 4$, and discount factor $\gamma = 0.95$. For each state-action pair, the next-state transitions are defined over four possible next states, with transition probabilities sampled from a Dirichlet distribution, $\text{Dir}(1,1,1,1)$. Among the 49 states excluding the fixed initial state, three are randomly selected as goal states, and each is assigned a distinct one-hot reward vector in the same manner as in the MO-Four-Room environment. To simulate the offline RL setting, a dataset of 100 trajectories is collected using a behavior policy with an optimality level of 0.5. Here, optimality is defined as the normalized performance relative to a uniformly random policy $\pi_{\text{unif}}$ (optimality = 0.0) and an optimal policy $\pi^{*}$ (optimality = 1.0). This implies that the behavior policy achieves performance halfway between that of the optimal and random policies. The experiments are repeated over 1000 seeds, with $\alpha\in\{0.0, 0.5, 1.0, 1.25\}$ and $\beta\in\{0.0001,0.0005, 0.001, 0.005, 0.01, 0.05, 0.1, 0.5, 1.0, 5.0, 10.0, 50.0, 100.0\}$, to provide a comprehensive analysis of the trade-off between welfare maximization and distributional shift.
\subsection{Additional visulaization: MO-Four-Room}
\label{Appendix_D_2}
Figure~\ref{fig1:Fourroom} illustrates that FairDICE maximizes the welfare objective, resulting in behavior that is distinct from conventional MORL approaches. In this subsection, we extend the experiment from Section~\ref{subsection_6.2} to further visualize two key insights proposed in our paper: (1) $\mu$ corresponds to the preference weights used in regularized Linear MORL, and (2) these weights are implicitly optimized to maximize the welfare objective—any deviation from the optimal $\mu$ leads to a reduction in overall welfare. In Section~\ref{subsection_6.2}, all preference weights were perturbed within Random MOMDP setting. While this effectively demonstrated that FairDICE selects the welfare-maximizing weights, it is not easy to visualize the consequence of deviation.

Given the optimal preference weight of FairDICE within MO-Four-Room, $\mathbf{\mu}^{*} = [\mu_1, \mu_2, \mu_3]$, we perturb $\mu_2$ and $\mu_3$ while keeping $\mu_1$ fixed. The perturbed weights are applied to FairDICE-fixed to obtain the corresponding optimal policy, whose performance is then evaluated as shown above.
\begin{figure}
  \centering
  \includegraphics[width=\linewidth]{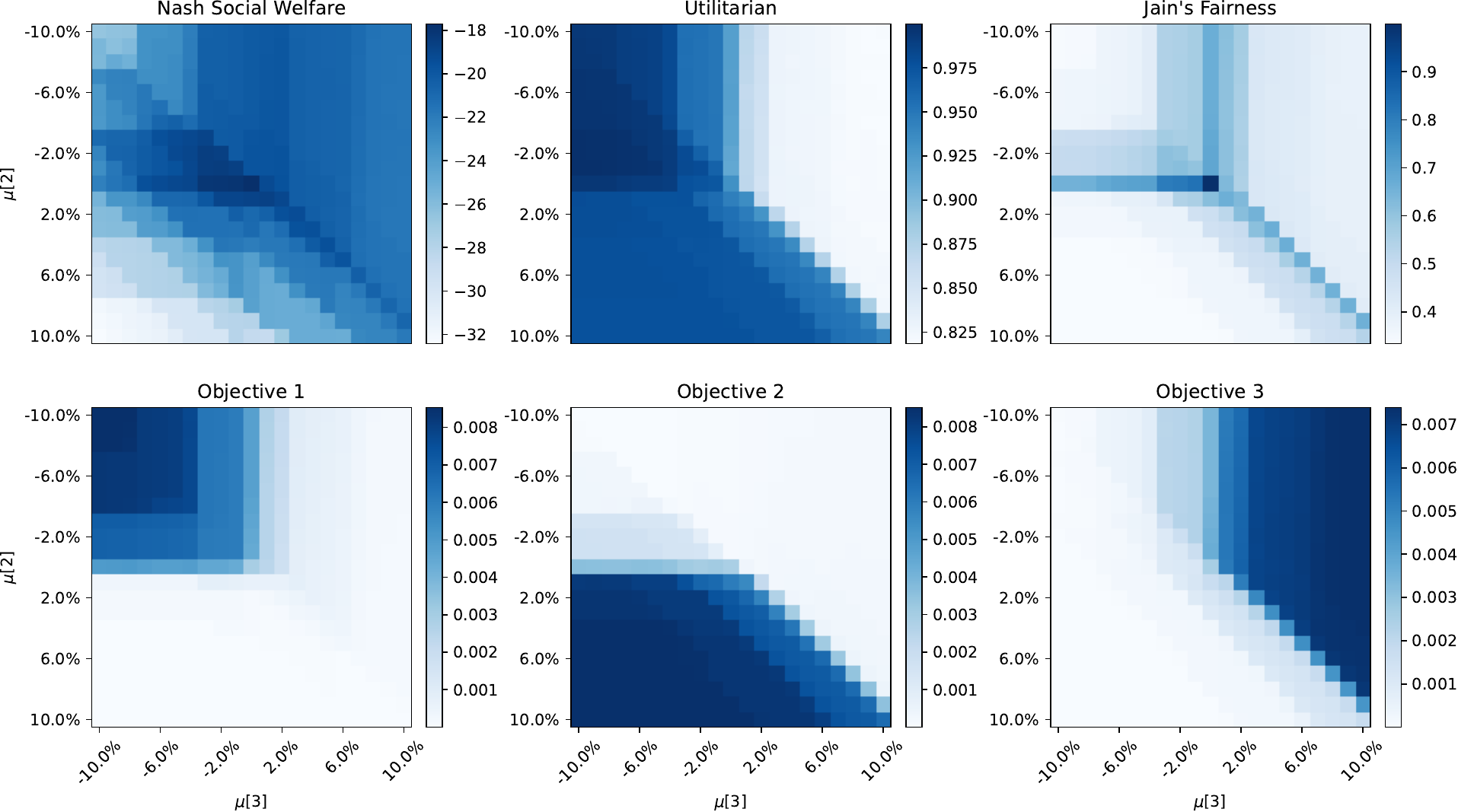}
  \caption{Visualization of FairDICE-fixed performance with different $\mu$ in MO-Four-Room. The center point is the optimal $\mu^{*}$ obtained by FairDICE. The x-axis and y-axis represent the degrees of perturbation applied to $\mu_2$ and $\mu_3$ from their optimal values.
  }
\end{figure}

The center point corresponds to the optimal solution of FairDICE, achieving the highest Nash social welfare. Increasing $\mu_2$ boosts return on objective 2, while increasing $\mu_3$ improves return on objective 3. Return on objective 1 increases when both $\mu_2$ and $\mu_3$ decrease. The point that maximizes Utilitarian welfare shifts toward prioritizing objective 1, but this comes at the cost of reduced Nash social welfare and lower Jain’s fairness index. These results highlight that FairDICE implicitly optimizes preference weights to maximize welfare, enabling fair behavior in the offline MORL setting.

\section{D4MORL Benchmark}
\label{appendix:d4morl}
To evaluate the efficacy of offline multi-objective reinforcement learning (MORL) algorithms, we utilize the D4MORL benchmark (with MIT License) introduced by \cite{zhu2023scaling}. D4MORL is the large-scale benchmark designed for offline MORL and includes high-dimensional continuous control environments derived from MuJoCo. We adopted settings and configurations of D4MORL without additional modifications. 
\subsection{Environments and Objectives}
D4MORL comprises six environments: MO-Ant, MO-HalfCheetah, MO-Hopper, MO-Swimmer, and MO-Walker2d, each with two conflicting objectives (e.g., speed vs. energy efficiency), and MO-Hopper-3obj, which includes three objectives—making it a more challenging benchmark. Each environment is defined by multiple, often conflicting, objectives such as forward velocity, jumping stability, or energy consumption. These objectives induce trade-offs, thereby enabling the study of Pareto-optimal policy learning in continuous control settings.

We summarize the objectives for each environment in Table~\ref{env-objectives}, including their physical interpretations and reward formulations. The rewards are computed based on physical quantities such as displacement, height, and control cost. Most environments include a survival bonus term $r_s$ and penalize excessive actions via an action cost term $r_a = \sum_k a_k^2$. The time delta $\Delta t$ determines the resolution of velocity and height-based rewards and is environment-specific.

\vspace{0.5em}
\noindent\textbf{Reward Terms and Environment-Specific Constants.}
The reward functions in Table~\ref{env-objectives} include shared terms whose values vary across environments:

\begin{itemize}
    \item \textbf{Action penalty} ($r_a$): Typically defined as the squared sum of action magnitudes, i.e., $r_a = \sum_k a_k^2$. However, different environments apply distinct scaling factors: $r_a = 0.5 \sum_k a_k^2$ in MO-Ant; $r_a = 2 \times 10^{-4} \sum_k a_k^2$ in MO-Hopper.
    
    \item \textbf{Survival bonus} ($r_s$): A constant reward to encourage survival, set to $r_s = 1.0$ in all environments except MO-Swimmer, where it is omitted.

    \item \textbf{Time delta} ($\Delta t$): Represents the duration between timesteps used in computing velocities or other dynamic terms. Its value is $\Delta t = 0.05$ in MO-Ant, MO-HalfCheetah, and MO-Swimmer; $\Delta t = 0.01$ in MO-Hopper and MO-Hopper-3obj; and $\Delta t = 0.008$ in MO-Walker2d.

    \item \textbf{Initial height} ($h_\text{init}$): In MO-Hopper and MO-Hopper-3obj, vertical jump rewards are defined relative to a fixed starting height of $h_\text{init} = 1.25$.
\end{itemize}

These constants follow the environment configurations detailed in Appendix~A of \cite{zhu2023scaling}.

\begin{table}[h]
  \caption{Objectives in D4MORL Environments}
  \label{env-objectives}
  \centering
  \begin{tabular}{lll}
    \toprule
    Environment & Objective Name & Reward Description \\
    \midrule
    MO-Ant & 
    \begin{tabular}[t]{@{}l@{}}
    $r_{vx}$: velocity in $x$ direction \\
    $r_{vy}$: velocity in $y$ direction
    \end{tabular}
    &
    \begin{tabular}[t]{@{}l@{}}
    $r_{vx} = \frac{x_t - x_{t-1}}{\Delta t} + r_s - r_a$ \\
    $r_{vy} = \frac{y_t - y_{t-1}}{\Delta t} + r_s - r_a$
    \end{tabular} \\
    \midrule
    MO-HalfCheetah &
    \begin{tabular}[t]{@{}l@{}}
    $r_v$: forward speed \\
    $r_e$: energy efficiency
    \end{tabular}
    &
    \begin{tabular}[t]{@{}l@{}}
    $r_v = \min(4.0, \frac{x_t - x_{t-1}}{\Delta t}) + r_s$ \\
    $r_e = 4.0 - r_a + r_s$
    \end{tabular} \\
    \midrule
    MO-Hopper &
    \begin{tabular}[t]{@{}l@{}}
    $r_r$: forward running \\
    $r_j$: vertical jumping
    \end{tabular}
    &
    \begin{tabular}[t]{@{}l@{}}
    $r_r = 1.5 \cdot \frac{x_t - x_{t-1}}{\Delta t} + r_s - r_a$ \\
    $r_j = 12 \cdot \frac{h_t - h_{init}}{\Delta t} + r_s - r_a$
    \end{tabular} \\
    \midrule
    MO-Hopper-3obj &
    \begin{tabular}[t]{@{}l@{}}
    $r_r$: forward running \\
    $r_j$: vertical jumping \\
    $r_e$: energy efficiency
    \end{tabular}
    &
    \begin{tabular}[t]{@{}l@{}}
    $r_r = 1.5 \cdot \frac{x_t - x_{t-1}}{\Delta t} + r_s$ \\
    $r_j = 12 \cdot \frac{h_t - h_{init}}{\Delta t} + r_s$ \\
    $r_e = 4.0 - r_a + r_s$
    \end{tabular} \\
    \midrule
    MO-Swimmer &
    \begin{tabular}[t]{@{}l@{}}
    $r_v$: forward speed \\
    $r_e$: energy efficiency
    \end{tabular}
    &
    \begin{tabular}[t]{@{}l@{}}
    $r_v = \frac{x_t - x_{t-1}}{\Delta t}$ \\
    $r_e = 0.3 - 0.15 \cdot r_a$
    \end{tabular} \\
    \midrule
    MO-Walker2d &
    \begin{tabular}[t]{@{}l@{}}
    $r_v$: forward speed \\
    $r_e$: energy efficiency
    \end{tabular}
    &
    \begin{tabular}[t]{@{}l@{}}
    $r_v = \frac{x_t - x_{t-1}}{\Delta t} + r_s$ \\
    $r_e = 4.0 - r_a + r_s$
    \end{tabular} \\
    \bottomrule
  \end{tabular}
\end{table}

\subsection{Behavioral Policy Quality}
Each dataset in D4MORL is collected using:
\begin{itemize}
    \item \textbf{Expert policy}: Selected from a large PGMORL\cite{xu2020prediction}-trained ensemble to match the target preference closely.
    \item \textbf{Amateur policy}: \textbf{Amateur policy:} Perturbed version of the expert policy, where actions have a certain probability of being stochastic. For most environments, stochastic actions are generated by scaling the expert action by a random factor sampled from $\text{Unif}(0.35, 1.65)$ (65\%), while the expert action is retained otherwise (35\%). In the MO-Swimmer environment, stochastic actions (35\%) are instead uniformly sampled from the action space to better approximate amateur-level performance.
\end{itemize}
The target preferences used during data collection are uniformly sampled from the full $(n-1)$-dimensional simplex. This promotes diverse trade-offs across objectives and ensures that the dataset covers a wide range of possible preferences. All preference vectors $\omega \in \mathbb{R}^n$ are normalized to satisfy $\omega_i \geq 0$ and $\sum_i \omega_i = 1$.

\subsection{Reward and State Normalization}
We introduce how rewards and states are normalized in the D4MORL benchmark \cite{zhu2023scaling}.

\textbf{Reward normalization.}
The reward values are normalized for each objective to the $[0, 1]$ range using min-max normalization:
\[
r^{\text{norm}} = \frac{r - r_{\min}}{r_{\max} - r_{\min}},
\]
where $r_{\min}, r_{\max}$ are computed empirically from the offline dataset for each objective dimension.

\textbf{State normalization.}
All state vectors are standardized using environment-specific statistics provided by the D4MORL benchmark. Specifically, the raw state $s$ is normalized as:
\[
s^{\text{norm}} = \frac{s - \mu}{\sigma},
\]
where $\mu$ and $\sigma$ denote the per-dimension mean and standard deviation of the state distribution, computed from the offline dataset.

\section{Implementation Details}
\label{appendix:implementation}

We use the Soft-$\chi^2$ divergence as the regularization function $f$ in FairDICE. This function is defined piecewise as:
\[
f(x) = 
\begin{cases}
x \log x - x + 1 & \text{if } x < 1 \\
\frac{1}{2}(x - 1)^2 & \text{otherwise}
\end{cases}
\]
This divergence combines the smooth behavior of KL divergence near $x = 0$ with the quadratic growth of the standard $\chi^2$ divergence for larger $x$.

Although the definition of the convex conjugate $f^*(y)$ may suggest a bi-level optimization, once $f(x)$ is specified, both $f^*(y)$ and $(f')^{-1}(x)$ can be obtained in closed form. For the chosen Soft-$\chi^2$ divergence, we have:
\[
f^*(y) =
\begin{cases}
e^{y} - 1, & y < 0, \\
\frac{1}{2}y^2 + y, & \text{otherwise},
\end{cases}
\quad
(f')^{-1}(x) =
\begin{cases}
e^{x}, & x < 0, \\
1 + x, & \text{otherwise}.
\end{cases}
\]
Therefore, the final FairDICE objective can be computed directly without solving any inner maximization loop.

\begin{algorithm}[htbp]
\caption{FairDICE}
\label{alg:fairdice}
\textbf{Input:} Offline dataset $D$, initial state distribution $p_0$, policy $\pi_\theta$, dual parameters $\nu_\psi$, $\mu_i$, divergence regulrarization parameter $\beta$, concave scalarization function $u_i$. \\
\textbf{Output:} Welfare-maximizing policy $\pi^*_\theta$
\begin{algorithmic}[1]
\State Initialize all parameters
\While{not converged}
    \State Update $\nu_\psi$, $\mu$ to minimize:
    \[
    \mathcal{L}_{\nu_\psi, \mu} = \mathbb{E}_{s \sim p_0}[(1 - \gamma)\nu_\psi(s)] 
    + \mathbb{E}_{(s, a) \sim D} \left[\beta f_0^*\left( \frac{e_{\nu_\psi, \mu}(s, a)}{\beta} \right)\right] 
    + \sum_i u^*_i(-\mu_i)
    \]
    \State Compute optimal weights: 
    \[
    w^*_{\nu_\psi, \mu}(s, a) = \max\left(0, (f')^{-1} \left( \frac{e_{\nu_\psi, \mu}(s, a)}{\beta} \right) \right)
    \]
    \State Update policy $\pi_\theta$ via weighted behavior cloning:
    \[
    \mathcal{L}_{\theta} = -\mathbb{E}_{(s, a) \sim D} \left[w^*_{\nu_\psi, \mu}(s, a) \cdot \log \pi_\theta(a \mid s)\right]
    \]
\EndWhile
\end{algorithmic}
\end{algorithm}

The FairDICE algorithm, summarized in Algorithm~\ref{alg:fairdice}, alternates between optimizing the dual variables $(\nu, \mu)$ and updating the policy $\pi$ via weighted behavior cloning. Both the policy $\pi$ and critic $\nu$ networks are implemented as multilayer perceptrons, parameterized by $\psi$ and $\theta$, respectively. The scalar parameters $\mu$ are updated to maximize the desired social welfare function, and we fix $\alpha = 1$ to correspond to the Nash social welfare objective. The initial state distribution $p_0$ is estimated from the offline dataset.

\renewcommand{\arraystretch}{1.4}
\begin{table}[h]
  \caption{Implementation Details for MO Environments}
  \label{tab:impl-details}
  \centering
  \begin{tabular}{ll}
    \toprule
    Hyperparameter & Value \\
    \midrule
    $\beta$ & \{1.0, 0.1, 0.01, 0.001, 0.0001\} \\
    Hidden dim of $\nu_\psi$ and $\pi_\theta$ & 768 (512 for MO-Ant) \\
    n\_layer of $\nu_\psi$ and $\pi_\theta$ & 3 (4 for MO-Hopper-3obj) \\
    Learning rate & $3 \times 10^{-4}$ \\
    $\gamma$ (discount factor) & 0.99 \\
    Optimizer & Adam \\
    \bottomrule
  \end{tabular}
  \vspace{0.5em}
  \caption*{\footnotesize Values in parentheses indicate environment-specific overrides.}
\end{table}
\renewcommand{\arraystretch}{1.0} 

Table~\ref{tab:impl-details} provides a summary of our default hyperparameters. The policy and value networks are constructed with three hidden layers, each containing 768 units. Optimization is performed using the Adam optimizer with a learning rate of $3 \times 10^{-4}$ and a discount factor of $\gamma = 0.99$. To study the effect of the regularization coefficient $\beta$—which governs the trade-off between distributional robustness and optimization stability—we conduct a hyperparameter sweep over $\beta \in \{1.0, 0.1, 0.01, 0.001, 0.0001\}$.
Our code is available at: \url{https://github.com/ku-dmlab/FairDICE.git}.

\section{Experiments Compute Resources}
\label{appendix:Resources}
All experiments were conducted on a single machine equipped with an Intel\textsuperscript{\textregistered} Xeon\textsuperscript{\textregistered} Gold 6330 CPU (256GB RAM) and an NVIDIA RTX 3090 GPU. Training a single FairDICE policy on each D4MORL task required approximately 10 to 20 minutes on average. During training, GPU memory usage remained below 20GB.

\section{Robustness of FairDICE to Limited Data Quality and Coverage in Offline Multi-Objective RL}
\label{appendix:robustness}

As offline RL methods are often sensitive to the quality and coverage of the dataset, we provide empirical evidence that FairDICE exhibits a degree of robustness to suboptimal data quality and limited coverage. Regarding data quality, we evaluated FairDICE on both the expert and amateur datasets from the D4MORL benchmark, and it consistently achieved high Nash Social Welfare across both settings. To further assess robustness to limited data coverage, we conducted additional experiments where we filtered out trajectories whose preference weights lie near the center of the simplex. Specifically, we removed trajectories in which all preference weights fall between 0.4 and 0.6. This filtering removes data points likely to represent balanced or fair trade-offs, resulting in a more challenging offline dataset.

\begin{table}[t]
\centering
\small 
\caption{Comparison of Nash Social Welfare before and after trajectory filtering across different environments and dataset qualities.}
\label{tab:nash-social-welfare}
\begin{tabular}{lccrr}
\toprule
\textbf{Environment} & \textbf{Dataset Quality} & \textbf{\% Traj. Removed} & \textbf{NSW (Full)} & \textbf{NSW (Filtered)} \\
\midrule
MO-Swimmer-v2       & Expert   & 24.0\% & 11.597 $\pm$ 0.091 & 11.489 $\pm$ 0.192 \\
MO-Swimmer-v2       & Amateur  & 24.0\% & 11.820 $\pm$ 0.005 & 11.819 $\pm$ 0.001 \\
MO-Walker2d-v2      & Expert   & 34.9\% & 11.534 $\pm$ 0.039 &  9.562 $\pm$ 1.161 \\
MO-Walker2d-v2      & Amateur  & 35.0\% & 11.396 $\pm$ 0.291 & 11.339 $\pm$ 0.016 \\
MO-Ant-v2           & Expert   & 43.1\% & 11.535 $\pm$ 0.018 & 11.320 $\pm$ 0.332 \\
MO-Ant-v2           & Amateur  & 43.2\% & 11.509 $\pm$ 0.049 & 11.384 $\pm$ 0.027 \\
MO-HalfCheetah-v2   & Expert   & 43.6\% & 11.828 $\pm$ 0.017 & 11.714 $\pm$ 0.035 \\
MO-HalfCheetah-v2   & Amateur  & 43.9\% & 11.994 $\pm$ 0.146 & 11.709 $\pm$ 0.074 \\
MO-Hopper-v2        & Expert   & 67.3\% & 11.058 $\pm$ 0.395 & 11.157 $\pm$ 0.014 \\
MO-Hopper-v2        & Amateur  & 67.7\% & 11.570 $\pm$ 0.003 & 11.548 $\pm$ 0.003 \\
\bottomrule
\end{tabular}
\end{table}

The results are shown in \ref{tab:nash-social-welfare}. Nash Social Welfare (Full) refers to performance on the original dataset without any filtering, while Nash Social Welfare (Filtered) reports performance after removing the trajectories. Each result shows the average Nash Social Welfare (NSW) over 5 seeds. FairDICE continued to perform reliably, demonstrating its ability to optimize fairness-driven objectives even under biased or sparse data conditions.

\section{Impact of f-divergence on FairDICE}
\label{appendix:alphaablation}
\begin{figure}[t]
  \centering
  \includegraphics[width=0.99\linewidth]{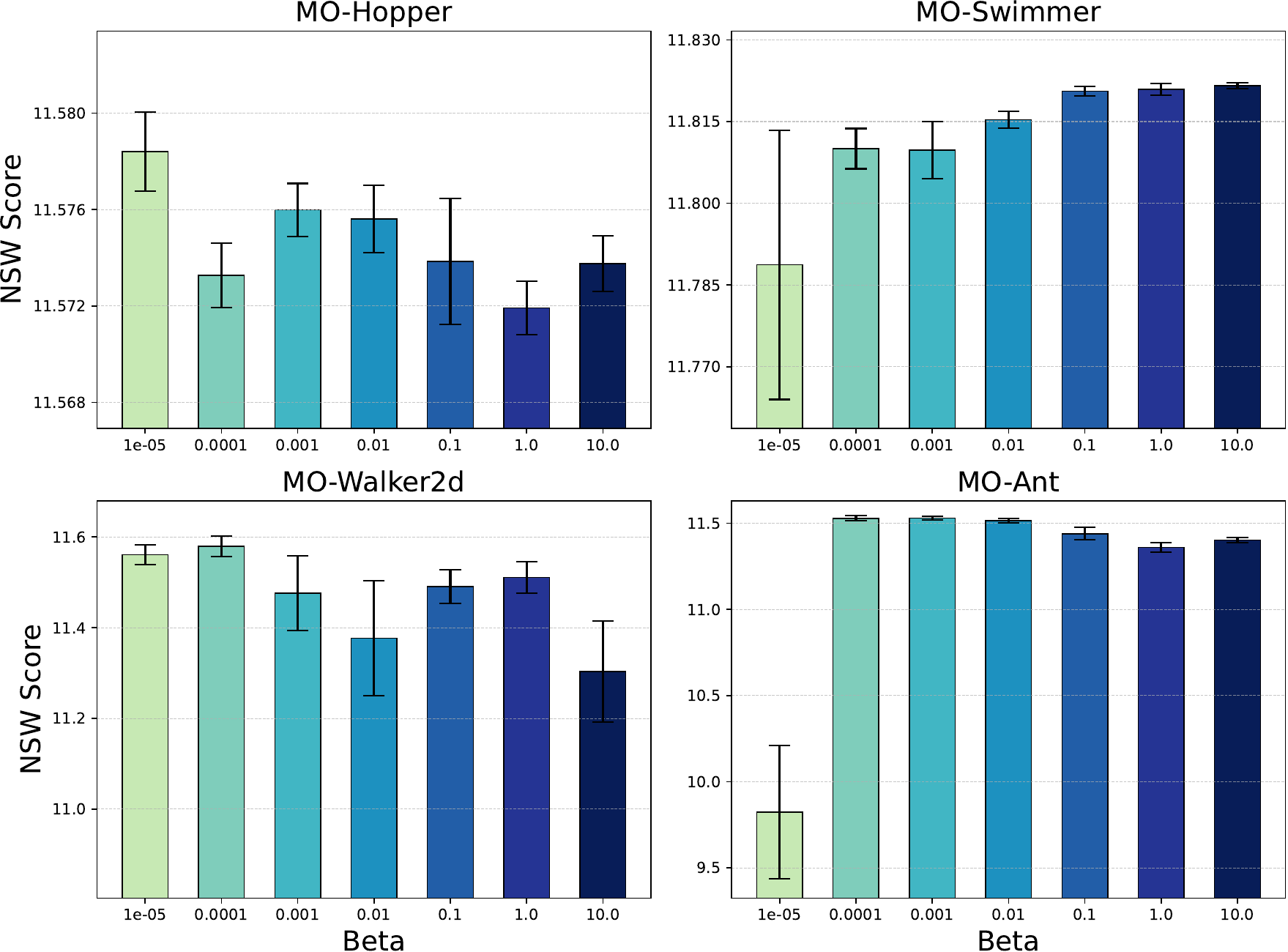}
  \caption{Performance of FairDICE with varying $\beta$ values on amateur datasets in D4MORL. Results are averaged over 10 seeds and 10 episodes, with error bars denoting ±1 standard errors.}
  \label{fig_beta}
  \vspace{-0.7cm}
\end{figure}
In this section, we investigate how varying $\beta$, which controls the strength of regularization, affects FairDICE’s performance. Figure~\ref{fig_beta} reports Nash social welfare (NSW) performance for $\beta \in \{10.0, 1.0, 0.1, 0.01, 0.001, 0.0001, 0.00001\}$. We show that the trade-off between NSW and distributional shift observed in FairDICE in the finite domain (Figure~\ref{fig2}) generally extends to the continuous domain. NSW performance typically improves as $\beta$ decreases, reflecting a stronger emphasis on maximizing NSW and reduced reliance on the dataset distribution. While FairDICE displays strong NSW across a wide range of $\beta$, excessive distributional shift at very small $\beta$ can degrade its practical performance.

An exception is observed in the MO-Swimmer environment, where NSW consistently decreases as $\beta$ decreases. This is likely because the MO-Swimmer dataset already contains trajectories with high NSW, making further deviation harmful. This is supported by Figure~\ref{fig:nsw_score_dist}, which shows that although trajectories were generated using different preference weights in existing preference-based baselines, they consistently achieve high NSW.

\end{document}